\documentclass[11pt,a4paper]{article}
\usepackage[hyperref]{emnlp-ijcnlp-2019}
\PassOptionsToPackage{sort&compress}{natbib}
\usepackage{times, latexsym, blindtext, amsmath, url, graphicx}
\usepackage{booktabs}
\usepackage{enumitem}
\usepackage{subcaption}
\usepackage{multirow}
\usepackage{tabularx}

\newcommand{\para}[1]{\noindent{\bf #1}}
\newcommand{\figref}[1]{Figure \ref{#1}}

\newcommand{\secref}[1]{Section \ref{#1}}
\newcommand{\tableref}[1]{Table \ref{#1}}

\newcommand{\lemma}[1]{\mathcal{V}_{\text{#1}}}
\newcommand{\cmv}{{\tt\small /r/ChangeMyView}\xspace}
\newcommand{\oppc}[1]{{\em \textcolor{purple}{#1}}\xspace}
\newcommand{\opword}[1]{{\em \textcolor{red}{#1}}\xspace}
\newcommand{\pcword}[1]{{\em \textcolor{blue}{#1}}\xspace}

\aclfinalcopy %

\title{What Gets Echoed? Understanding the ``Pointers'' in Explanations of Persuasive Arguments}

\author{David Atkinson \and Kumar Bhargav Srinivasan \and Chenhao Tan\\
  Department of Computer Science \\
  University of Colorado Boulder \\
  Boulder, CO \\
  {\tt david.i.atkinson, kumar.srinivasan, chenhao.tan@colorado.edu} \\
}

\date{}

\begin{document}
\maketitle
\begin{abstract}
Explanations are central to everyday life, and are 
a
topic of growing interest in the
AI community. To investigate the process of providing natural language explanations, we leverage the dynamics of the \cmv subreddit to build a dataset with 36K naturally occurring explanations of why an argument is persuasive. We propose a novel word-level prediction task to investigate how explanations selectively reuse, or \emph{echo}, information from what is being explained (henceforth, \emph{explanandum}). 
We develop features to capture the properties of a word in the explanandum, and show that our proposed features 
not only have 
relatively 
strong predictive power 
on the echoing of a word in an explanation, but also enhance neural methods of generating explanations. In particular, 
while the non-contextual properties of a word itself are more valuable for stopwords,
the interaction between the constituent parts of an explanandum is crucial in predicting the echoing of content words. We also find intriguing patterns of a word being echoed. For example, although nouns are generally less likely to be echoed, subjects and objects can, depending on their source, be more likely to be echoed in the explanations.

\end{abstract}

\section{Introduction}
\label{sec:intro}

Explanations are essential for understanding and learning \citep{keil2006explanation}.
They can take many forms, ranging from everyday explanations for questions such as why one likes Star Wars, to sophisticated formalization 
in the philosophy of science \citep{salmon2006four}, to simply highlighting features in recent work on interpretable machine learning \citep{ribeiro2016should}.

\begin{table*}
\small
\centering
\begin{tabular}{p{0.95\textwidth}}
\toprule
{\bf Original post (OP):} CMV: most hit music artists today are bad musicians \\
Now I know, music is art and art has no rules, but this is only so true. Movies are art too but I think most of us can agree the emoji movie was objectively bad. That aside: I really feel like once you remove the persona and performances of the artists from the "top 40" songs and listen to them as just a song, most are objectively bad. They're super repetitive, the lyrics and painfully generic, and there's hardly ever anything new or challenging. And from what I understand most of these artists don't even write their own songs. Of course there are exceptions but I find them to be extremely rare. It seems to me they're only popular become of who they are and how they look/perform. I realize this is probably a very snobbish view which is why I want to be enlightened, so can anyone convince me otherwise? Are they actually good musicians or just good performers? [one more paragraph ...]\\
\midrule
{\bf Persuasive comment (PC):} Music appreciation is a skill, and it's all about pattern recognition.\\
When we're children, we need songs that are really simple, repetitive and with easy to recognize patterns. The younger we are, the simpler the songs. Toddlers like nursery rhymes, lullabies, jingles. Teens like pop music. And teens spend more on music than anyone else. [four more paragraphs ...] \\
Lastly, you have to consider that music can be listened to in different ways and for different purposes. You can listen to it alone on headphones, and think about what it means and how it makes you feel. Or you can dance to it with your friends. Or maybe you need something on in the background during a dinner party, or a house party, or while you study, or are trying to fall asleep, or work out. Pop music is really good in some of these situations, really bad in others. But it serves a definite purpose and isn't bad in any essential way.\\
\midrule
{\bf Explanation:}
$\Delta$ I guess I never \oppc{really} \opword{looked} at it as \oppc{music} \pcword{serving} \pcword{different} \pcword{purposes}. I can \opword{see} how \pcword{pop} \oppc{music} fills a certain \pcword{purpose}, and I guess the \opword{artist} does \oppc{n't} necessarily have to be the \pcword{one} to \opword{write} the \oppc{song} (although I \pcword{appreciate} it when they do).\\
\bottomrule
\end{tabular}
\caption{An illustration of the pointers in an example explanation of \cmv.
We color the words in the explanation based on whether it is used in the original post (e.g., \opword{artist}), in the persuasive comment (e.g., \pcword{purpose}), or both (e.g., \oppc{music}).
We stem all the words before matching and do not color stopwords for readability.
}
\label{tb:example}
\end{table*}

Although everyday explanations are mostly encoded in natural language,
natural language explanations remain understudied in NLP, partly due to a lack of appropriate datasets and problem formulations.
To address these challenges,
we 
leverage \cmv, a community dedicated to sharing counterarguments to controversial views on Reddit, to build a sizable dataset of naturally-occurring explanations.
Specifically, in \cmv, an original poster (OP)
first delineates the rationales for a (controversial) opinion (e.g., in \tableref{tb:example}, ``most hit music artists today are bad musicians'').
Members of \cmv are invited to provide counterarguments.
If a counterargument changes the OP's view, the OP awards a $\Delta$ to indicate the change and is required to {\em explain why the counterargument is persuasive}.
In this work, we refer to what is being explained, including both the original post and the persuasive comment, as the {\em explanandum}.\footnote{The plural of {\em explanandum} is explananda.}

An important advantage of explanations in \cmv is that 
the explanandum 
contains most of the required information 
to provide its explanation.
These explanations often select key counterarguments in the persuasive comment and connect them with the original post.
As shown in \tableref{tb:example}, the explanation naturally points to, or {\em echoes}, part of the explanandum (including both the persuasive comment and the original post) 
and in this case highlights the argument of ``music serving different purposes.''

These naturally-occurring explanations thus enable us to computationally investigate 
the selective nature of explanations:
``people rarely, if ever, expect an explanation that consists of an actual and complete cause of an event. Humans are adept at selecting one or two causes from a sometimes infinite number of causes to be the explanation'' \citep{miller2018explanation}.
To understand the selective process of providing explanations, we formulate a word-level 
task to predict whether a word in 
an explanandum will be echoed in its explanation.

Inspired by the 
observation that words that are likely to be echoed are either frequent or rare, we propose a variety of features to capture how a word is used in the explanandum as well as 
its non-contextual properties in \secref{sec:features}.
We find that a word's usage in the original post and in the persuasive argument 
are 
similarly related to being echoed, except in part-of-speech tags and grammatical relations.
For instance, verbs in the original post are less likely to be echoed, while the relationship is reversed in the persuasive argument.

We further demonstrate that these features can significantly outperform a random baseline and 
even a neural model with significantly more knowledge of a word's context.
The difficulty of predicting 
whether content words 
 (i.e., non-stopwords)
are echoed
is much greater than that of stopwords,\footnote{We use the stopword list in NLTK.} among which adjectives are the most difficult and nouns are relatively the easiest.
This observation highlights the important role of nouns in 
explanations.
We also find that 
the relationship between a word's usage in the original post and in the persuasive comment is crucial for predicting the echoing of content words.
Our proposed features can also improve the performance of pointer generator networks 
with coverage 
in generating explanations \citep{see_get_2017}.

To summarize, our main contributions are:

\begin{itemize}[itemsep=0pt,leftmargin=*,topsep=0pt]
    \item We highlight the importance of computationally characterizing human explanations and formulate a concrete problem of predicting how information is selected from explananda to form explanations, including building a novel dataset of naturally-occurring explanations.
    \item We provide 
    a computational characterization of natural language explanations and demonstrate the U-shape in which words get echoed.
    \item We identify interesting patterns in what gets echoed through a novel word-level classification task, including the importance of nouns in shaping explanations and the importance of contextual properties of both the original post and persuasive comment in predicting the echoing of content words. 
    \item We show that vanilla LSTMs fail to learn some of the features we develop and that the proposed features can even improve performance in generating explanations with pointer networks.
\end{itemize}

Our code and dataset is available at \url{https://chenhaot.com/papers/explanation-pointers.html}.

\section{Related Work}
\label{sec:related}

To provide background for our study, we first present a brief overview of explanations for the NLP community, and then discuss the connection of our study with pointer networks, linguistic accommodation, and argumentation mining.

The most developed discussion of explanations is in the philosophy of science. 
Extensive studies aim to develop formal models of explanations (e.g., the deductive-nomological model in \citet{hempel1948studies}, see \citet{salmon2006four} and \citet{woodward2005making} for a review).
In this view, explanations are like proofs in logic.
On the other hand, psychology and cognitive sciences examine ``everyday explanations'' \citep{keil2006explanation,lombrozo2006structure}.
These explanations tend to be selective,
are typically encoded in natural language, and shape our understanding and learning in life despite the absence of ``axioms.''
Please refer to \citet{wilson_shadows_1998} for a detailed comparison of these two modes of explanation.
Although explanations have attracted significant interest from the AI community thanks to the growing interest on interpretable machine learning \citep{doshi2017towards,lipton2016mythos,guidotti2019survey},
such studies seldom refer to prior work in social sciences
\citep{miller2018explanation}.
Recent studies also show that explanations such as highlighting important features induce limited improvement on human performance in detecting deceptive reviews and media biases \citep{lai+tan:19,horne2019rating}.
Therefore, we believe that developing a computational understanding of everyday explanations is crucial for explainable AI.
Here we provide a data-driven study of everyday explanations
in the context of persuasion.

In particular, we investigate the ``pointers'' in explanations, inspired by recent work on pointer networks \citep{vinyals2015pointer}.
Copying mechanisms allow a decoder to generate a token by copying from the source, and have been shown 
to be effective
in generation tasks ranging from summarization to program synthesis \citep{see_get_2017,ling-etal-2016-latent,gu_incorporating_2016}.
To the best of our knowledge, our work is the first 
to
investigate the phenomenon of pointers in explanations.

Linguistic accommodation and studies on quotations also examine the phenomenon of reusing words \citep{Danescu-Niculescu-Mizil:2011:MMW:1963405.1963509,giles2007communication,leskovec2009meme,simmons2011memes}.
For instance, \citet{echoes} show that power differences are reflected in 
the echoing of function words; \citet{chenhaopres} find that news media prefer to quote locally distinct sentences in political debates.
In comparison, our word-level formulation presents a 
fine-grained view of echoing words, and 
puts a stronger emphasis on
content words than 
work on linguistic accommodation.

Finally, our work is concerned with an especially challenging problem in social interaction: persuasion.
A battery of studies have done work to enhance our understanding of persuasive arguments \citep{wang2017winning,justine,habernal2016makes,lukin2017argument,esin},
and the area of argumentation mining specifically investigates the structure of arguments \citep{lippi2016argumentation,walker2012stance,somasundaran2009recognizing}.
We build on previous work by \citet{tan_winning_2016} and leverage the 
dynamics of \cmv.
Although our findings are certainly related to the persuasion process, we focus on understanding the self-described reasons for persuasion, instead of the structure of arguments or the factors that drive effective persuasion.

\section{Dataset}
\label{sec:dataset}

\begin{figure*}[t]
\centering
\begin{subfigure}[t]{0.32\textwidth}
  \centering
  \includegraphics[width=\textwidth]{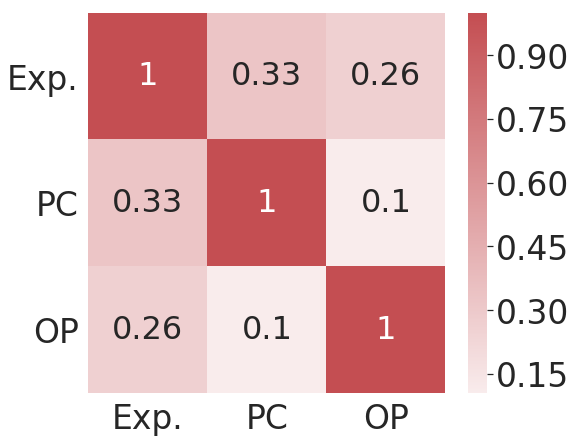}
  \caption{Length correlations.}
  \label{fig:lengths_heatmap}
\end{subfigure}
\hfill
\begin{subfigure}[t]{0.32\textwidth}
  \centering
  \includegraphics[width=0.83\textwidth]{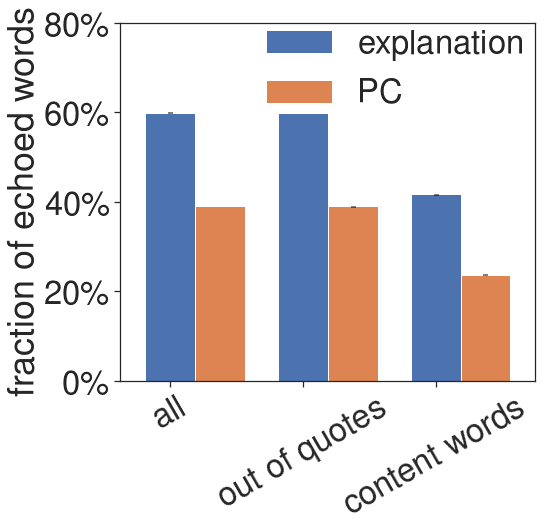}
  \caption{Fraction of words that are echoed from the explanandum.}
  \label{fig:copy}
\end{subfigure}
\hfill
\begin{subfigure}[t]{0.32\textwidth}
  \centering
  \includegraphics[width=0.9\textwidth]{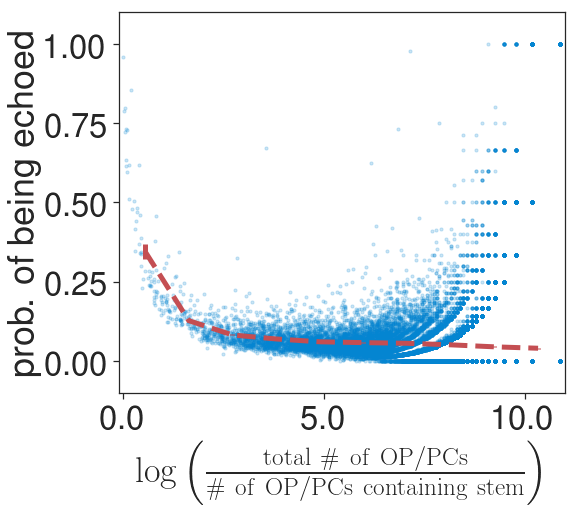}
  \caption{Word-level echoing probability vs. document frequency.}
  \label{fig:copy-oppc_to_exp}
\end{subfigure}
\caption{\figref{fig:lengths_heatmap} shows the pairwise Pearson correlation coefficient between lengths of OP, PC, and explanation (all values are statistically significant with $p<1\mathrm{e}{-10}$).
\figref{fig:copy} shows the average fraction of words in an explanation that are in its OP or PC, and the fraction of words in a PC that are in its OP.
In \figref{fig:copy-oppc_to_exp}, the $y$-axis represents the probability of a word in an OP or PC being echoed in the explanation, while the $x$-axis shows the inverse document frequency of that word in training data. For each document frequency decile, we calculate the probability of a word in that decile being echoed, and plot those probabilities with the red line. In \figref{fig:copy} and \figref{fig:copy-oppc_to_exp}, the (small) error bars represent standard errors.
}
\label{fig:dataset}
\end{figure*}

Our dataset is derived from the \cmv subreddit, which has more than 720K subscribers \cite{tan_winning_2016}.
\cmv hosts conversations where someone expresses a view and others then try to change that person's mind.
Despite being fundamentally based on argument, \cmv has a reputation for being remarkably civil and productive \cite{cmvmedia}, e.g., a journalist wrote ``In a culture of brittle talking points that we guard with our lives, Change My View is a source of motion and surprise'' \cite{heffernan:2018a}.

The delta mechanism in \cmv allows members to acknowledge opinion changes and enables us to identify {\em explanations} for opinion changes \citep{deltasystem}. Specifically, it requires
``Any user, whether they're the OP or not, should reply to a comment that changed their view with a delta symbol and {\em an explanation of the change}.''
As a result, we have access to tens of thousands of naturally-occurring explanations and associated explananda.
In this work, we focus on the opinion changes of the original posters.

Throughout this paper, we use the following terminology:

\begin{itemize}[itemsep=-5pt,leftmargin=*,topsep=0pt]
    \item An {\bf original post (OP)} is an initial post where the original poster justifies his or her opinion.
    We also use OP to refer to the original poster.
    \item A {\bf persuasive comment (PC)} is a comment that directly leads to an opinion change on the part of the OP (i.e., winning a $\Delta$).
    \item A {\bf top-level comment} is a comment that directly replies to an OP, and \cmv requires the top-level comment to ``challenge at least one aspect of OP’s stated view (however minor), unless they are asking a clarifying question.''
    \item An {\bf explanation} is a comment where an OP acknowledges a change in his or her view and provides an explanation of the change.
    As shown in \tableref{tb:example},
    the explanation not only provides a rationale, it can also include other discourse acts, such as expressing gratitude.
\end{itemize}

Using \url{https://pushshift.io}, we collect the posts and comments in 
\cmv 
from January 17th, 2013 to January 31st, 2019, and extract tuples of (OP, PC, explanation).
We use the tuples from the final six months of our dataset as the test set, those from the six months before that as the validation set, and the remaining tuples as the training set. The sets contain 5,270, 5,831, and 26,617 tuples respectively.
Note that there is no overlap in time between the three sets and the test set can therefore be used to assess generalization
including potential changes in community norms and world events.

\para{Preprocessing.} We perform a number of preprocessing steps, such as converting 
blockquotes in Markdown to quotes, filtering explicit edits made by authors, mapping all URLs to a special \texttt{@url@} token, and replacing hyperlinks with the link text. 
We ignore all triples that contain any deleted comments or posts.
We use spaCy for tokenization and tagging \citep{honnibal_spacy_2017}.
We also use the NLTK implementation of the Porter stemming algorithm to store the stemmed version of each word, for later use in our prediction task \citep{loper2002nltk, porter1980algorithm}. Refer to the supplementary material for more information on preprocessing.

\para{Data statistics.}
\tableref{tab:basic_stats} provides basic statistics of the training tuples and how they compare to other comments.
We highlight the fact that PCs 
are on average longer than top-level comments, suggesting that 
PCs contain substantial counterarguments that directly contribute to opinion change.
Therefore, we simplify the problem by focusing on the (OP, PC, explanation) tuples and ignore any other exchanges between an OP and a commenter.

Below, we highlight some notable features of explanations as they appear in our dataset.

\para{The length of explanations shows stronger correlation with that of OPs and PCs than between OPs and PCs (\figref{fig:lengths_heatmap}).}
This observation indicates that explanations are somehow better related with OPs and PCs than PCs are with OPs in terms of language use.
A possible reason is that the explainer combines their natural tendency towards length with accommodating the PC.

\begin{table}[t]
    \centering
    \small
    \begin{tabular}{p{2.4cm}rrr}
        \toprule
        & count & \#sentences & \#words\\
        \midrule
        \multicolumn{4}{c}{Tuples of (OP, PC, Explanations)}\\
        Original Posts & 26.3K & 16.8 & 298.8 \\ 
        Persuasive comments & 26.3K & 12.6 & 218.3 \\
        Explanations & 26.3K & 5.3 & 79.8  \\  
        \midrule
        \multicolumn{4}{c}{All of \cmv during the training period} \\
        Original posts  & 93.4k & 13.2 & 172.6 \\ 
        Top-level comments & 681.6k & 9.1 & 147.4 \\
        All comments  & 3.6M & 6.5 & 98.9 \\
        \bottomrule
    \end{tabular}
    \caption{Basic statistics of the training dataset.}
    \label{tab:basic_stats}
\end{table}

\para{Explanations have a greater fraction of ``pointers'' than do persuasive comments (\figref{fig:copy}).}
We measure the likelihood of a word in an explanation being copied from either its OP or PC and provide a similar probability for a PC for copying from its OP.
As we discussed in \secref{sec:intro}, the words in an explanation are much more likely to come from the existing discussion than are the words in a PC (59.8\% vs 39.0\%).
This phenomenon holds even if we restrict ourselves to considering words outside quotations, which removes the effect of quoting other parts of the discussion,
and if we focus only on content words, which removes the effect of ``reusing'' stopwords.

\para{Relation between a word being echoed and its document frequency (\figref{fig:copy-oppc_to_exp}).}
Finally, as a preview of our main results,
the document frequency of a word from the explanandum is related to the probability of being echoed in the explanation. Although the average likelihood declines as the document frequency gets lower, we observe an intriguing U-shape in the scatter plot.\footnote{A similar U-shape exists if we examine the probability of a PC echoing its OP, but does not show up if we compare an OP echoing a different, randomly chosen OP.
  It is worth noting that PCs can also be viewed as explaining why the OP is problematic. However, constructing a PC involves selecting from a large number of possible counter perspectives (all of which are unobservable).
See the supplementary material for a detailed discussion.
}
In other words, the words that are most likely to be echoed are either unusually frequent or unusually rare, while most words in the middle show a moderate likelihood of being echoed.

\begin{table*}[t!]
\centering
\small
\begin{tabular}{p{2cm}p{11cm}p{1.5cm}}
\toprule
Feature group & Features and intuitions & Echoed? \\
\midrule
\multirow{4}{2cm}{Non-contextual properties} & Inverse document frequency. As shown in \figref{fig:copy-oppc_to_exp}, although document frequency can have non-linear relationships with being copied, the average echoing probability is greater for more common words. & $\downarrow\downarrow\downarrow\downarrow$ \\
& Number of characters. Longer words tend to be more complicated, and may be more likely to be echoed as part of the core argument. & $\downarrow\downarrow\downarrow\downarrow$ \\
& Wordnet depth. Similar to number of characters, the depth in wordnet can indicate the complexity of a word and we expect words with greater depth to be echoed. & $\downarrow\downarrow\downarrow\downarrow^{RC,RS}$\\
& Echoing likelihood. We also compute the general tendency of a word being echoed in the training data. We expect the feature to be positively correlated with the label.& $\uparrow\uparrow\uparrow\uparrow$\\
\midrule
\multirow{5}{2cm}{How a word is used in an OP or PC (OP/PC usage)} & \multicolumn{2}{p{13cm}}{Part-of-speech (POS) tags. We compute the percentage of times that the surface forms of a stemmed word appear as different part-of-speech tags.
We expect nouns and verbs more likely to be echoed. Results: verb in an OP $\downarrow\downarrow\downarrow\downarrow^{RS}$, noun in an OP ($\downarrow\downarrow\downarrow\downarrow$), verb in a PC ($\uparrow\uparrow\uparrow\uparrow$), noun in a PC: $\downarrow\downarrow\downarrow\downarrow^{RC}$.
} 
\\
& \multicolumn{2}{p{13cm}}{Subjects and objects from dependency labels. We compute 
the percentage of times that the word appears as subjects, objects, and others.
We expect subjects and objects more likely to be echoed.
Results: subjects in an OP: $\uparrow\uparrow\uparrow\uparrow$, objects in an OP: $\downarrow\downarrow\downarrow\downarrow^{RC}$, others in an OP: $\uparrow\uparrow\uparrow\uparrow^{RC}$,
subjects in a PC: $\downarrow\downarrow\downarrow\downarrow$, objects in a PC: $\downarrow\downarrow\downarrow\downarrow$; others in a PC: $\uparrow\uparrow\uparrow\uparrow$.
} \\
& (Normalized) term frequency. We expect frequent terms to be echoed. & $\uparrow\uparrow\uparrow\uparrow$ \\
& \#surface forms.  We expect words that have diverse surface forms to be echoed. & $\uparrow\uparrow\uparrow\uparrow$ \\
& \multicolumn{2}{p{13cm}}{Location. For words that never show up in an OP or PC, the default value is 0.5. We expect later words to be echoed. Results: location in an OP: $\uparrow\uparrow\uparrow\uparrow$ (not significant in stopwords); location in a PC:  $\uparrow^{RS}$.} \\
& In quotes. We expect words in quotes to be echoed as they are already emphasized. & $\uparrow\uparrow\uparrow\uparrow$ \\
& Entity. We expect entities to be echoed. 
& $\uparrow\uparrow\uparrow\uparrow$\\
\midrule
\multirow{4}{2cm}{How a word connects an OP and PC (OP-PC relation)} & Occurs both in an OP and PC. &  $\uparrow\uparrow\uparrow\uparrow$ \\
& \#Surface forms in an OP but not in the PC.& $\downarrow\downarrow\downarrow\downarrow$ \\
& \#Surface forms in a PC but not in the OP.& $\uparrow\uparrow\uparrow\uparrow^{RS}$\\
& Jensen-Shannon (JS) distance between the OP and PC POS tag distributions of the word. & $\downarrow\downarrow\downarrow\downarrow$\\
& 
JS distance between subjects/objects distributions of the word in an OP and PC. & $\downarrow\downarrow\downarrow\downarrow$ \\
\midrule
\multirow{5}{2cm}{General OP/PC properties} 
& OP length. & $\downarrow\downarrow\downarrow\downarrow^{RS}$\\
& PC length. & $\uparrow\uparrow\uparrow\uparrow$\\
& Difference in \#words. & $\downarrow\downarrow\downarrow\downarrow^{RS}$ \\
& Difference in average \#characters in words. & $\downarrow\downarrow\downarrow\downarrow$\\
& Part-of-speech tags distributional differences between an OP and PC. &  $\downarrow\downarrow\downarrow\downarrow$ \\
& Depth of the PC in the thread. & $\uparrow\uparrow\uparrow\uparrow$ \\
\bottomrule
\end{tabular}
\vspace{-0.1in}
\caption{Features to capture the properties of a word in the context of an explanandum. The last column shows $t-$test results after Bonferroni correction. 
$\uparrow$ indicates that words that are echoed have a greater value in the feature, while $\downarrow$ indicates the reverse.
The number of arrows indicates the level of p-value: $\uparrow\uparrow\uparrow\uparrow$: $p<0.0001$, $\uparrow\uparrow\uparrow$: $p<0.001$, $\uparrow\uparrow$: $p<0.01$, $\uparrow$: $p<0.05$.
$^{RC}$ and $^{RS}$ indicate that the direction is flipped in content words and stopwords respectively. 
We show significance testing results in a condensed format for space reasons.
Refer to the supplementary material for the complete testing results.
}
\vspace{-0.1in}
\label{tb:features}
\end{table*}

\section{Understanding the Pointers in Explanations}
\label{sec:features}

To further investigate 
how explanations select words from the explanandum,
we formulate a word-level prediction task to predict whether words in an OP or PC are echoed in its explanation.
Formally, given a tuple of (OP, PC, explanation), we extract the unique stemmed words 
as $\lemma{OP}, \lemma{PC}, \lemma{EXP}$.
We then define the label for each word in the OP or PC, $w \in \lemma{OP} \cup \lemma{PC}$, based on the explanation as follows:
\begin{align*}
y_w = \begin{cases} 
1 &\mbox{if } w \in \lemma{EXP}, \\ 
0 & \mbox{otherwise. } \end{cases}
\end{align*}

\noindent Our prediction task is thus a straightforward binary classification task at the word level.
We develop the following five groups of features to capture properties of how a word is used in the explanandum (see \tableref{tb:features} for the full list):

\begin{itemize}[itemsep=0pt,leftmargin=*,topsep=0pt]
    \item Non-contextual properties of a word.
    These features are 
    derived 
    directly from the word and capture the general tendency of a word being echoed in explanations.
    \item Word usage in an OP or PC (two groups). These features capture \emph{how} a word is used in an OP or PC. As a result, for each feature, we have two values for the OP and PC respectively.
    \item How a word connects an OP and PC.
    These features look at the difference between word usage 
    in the OP and PC.
    We expect this group to be the most important in our task.
    \item General OP/PC properties. These features capture the general properties of a conversation.
    They can be used to characterize the background distribution of echoing.
\end{itemize}

\tableref{tb:features} further shows the intuition for including each feature, and condensed $t$-test results after Bonferroni correction.
Specifically, we test whether the words that were echoed in explanations have different feature values from those that were not echoed.
In addition to considering all words, we also separately consider stopwords and content words 
in light of \figref{fig:copy-oppc_to_exp}.
Here, we highlight a few observations:

\begin{itemize}[itemsep=0pt,leftmargin=*,topsep=0pt]
    \item Although we expect more complicated words ({\em \#characters}) to be echoed more often, this is not the case on average.
    We also observe an interesting example of Simpson's paradox in the results for Wordnet depth \citep{blyth1972simpson}:
    shallower words are more likely to be echoed across all words, but deeper words are more likely to be echoed 
    in content words and stopwords.
    \item OPs and PCs generally exhibit similar behavior for most features, except for part-of-speech and grammatical relation (subject, object, and other.)
    For instance, verbs in an OP are less likely to be echoed, while verbs in a PC are more likely to be echoed.

    \item Although nouns from both OPs and PCs are less likely to be echoed, within content words, subjects and objects from an OP 
    are more likely to be echoed. 
    Surprisingly, subjects and objects in a PC are less likely to be echoed, which suggests that the original poster tends to 
    refer back to their own subjects and objects, or introduce new ones, 
    when providing explanations.

    \item Later words in OPs and PCs are more likely to be echoed, especially in OPs.
    This could relate to OPs summarizing their rationales at the end of their post and PCs putting their strongest points last.

    \item Although the number of surface forms in an OP or PC is
    positively correlated with being echoed, the differences in surface forms show reverse trends:
    the more surface forms of a word that show up only in the PC (i.e., not in the OP), the more likely a word is to be echoed. However, the reverse is true for the number of surface forms in only the OP.
    Such contrast 
    echoes \citet{tan_winning_2016}, in which dissimilarity in word usage between the OP and PC was a predictive feature of successful persuasion.

\end{itemize}

\section{Predicting Pointers}
\label{sec:prediction}

We further examine the effectiveness of our proposed features in a predictive setting.
These features achieve strong performance in the word-level classification task, and can enhance neural models in both the word-level task and generating explanations.
However, the word-level task remains challenging, especially for content words.

\begin{figure*}[t]
\centering
\begin{subfigure}[t]{0.33\textwidth}
  \centering
  \includegraphics[width=\textwidth]{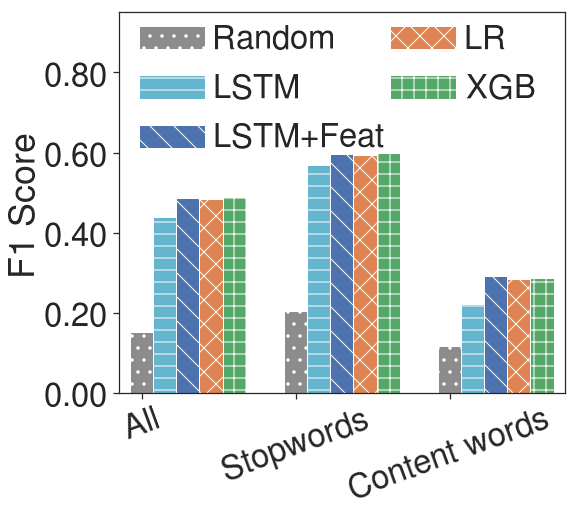}
  \caption{Overall Performance comparison between models.}
  \label{fig:exp_overall_perf}
\end{subfigure}
\hfill
\begin{subfigure}[t]{0.31\textwidth}
  \centering
  \includegraphics[width=\textwidth]{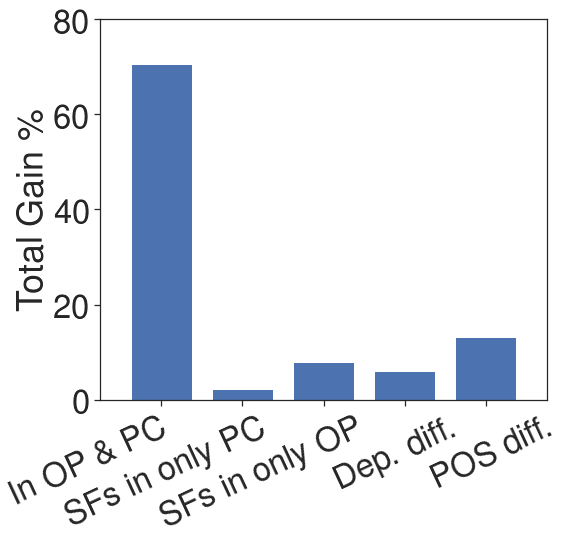}
  \caption{Feature importance of an ablated model with OP-PC relation.}
  \label{fig:exp_importance}
\end{subfigure}
\hfill
\begin{subfigure}[t]{0.33\textwidth}
  \centering
  \includegraphics[width=\textwidth]{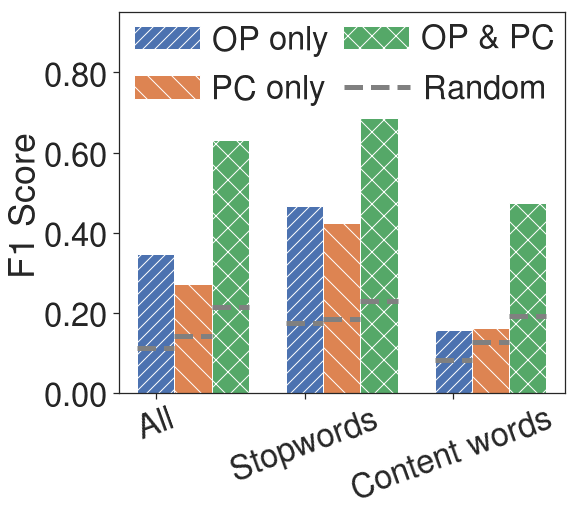}
  \caption{Performance vs. word source.}
  \label{fig:exp_oppc_perf}
\end{subfigure}
\caption{\figref{fig:exp_overall_perf} presents the performance of different models.
We evaluate the performance of each model on the subset of stopwords and content words.
Our features with XGBoost and logistic regression outperform the vanilla LSTM model, and adding our features to the vanilla LSTM model achieves similar performance as XGBoost.
\figref{fig:exp_importance} shows the normalized total gain of the classifier only based on features in OP-PC relation, while \figref{fig:exp_oppc_perf} further breaks down the performance based on where the words come from.
}
\label{fig:perf}
\end{figure*}

\subsection{Experiment setup}

We consider two classifiers for our word-level classification task:
logistic regression and gradient boosting tree (XGBoost) \citep{chen2016xgboost}.
We hypothesized that XGBoost would outperform logistic regression because our problem is non-linear, as shown in \figref{fig:copy-oppc_to_exp}.
   
To examine the utility of our features in a neural framework, we further adapt our word-level task as a tagging task, and use LSTM as a baseline.
Specifically, we concatenate an OP and PC with a special token as the separator so that an LSTM model can potentially distinguish the OP from PC, and then tag each word based on the label of its stemmed version.
We use GloVe embeddings to initialize the word embeddings \cite{pennington_glove:_2014}.
We concatenate our proposed features of the corresponding stemmed word to the word embedding;
the resulting difference in performance between
a vanilla LSTM demonstrates the utility of our proposed features.
We scale all features to $[0, 1]$ before fitting the models.
As introduced in \secref{sec:dataset}, we split our tuples of (OP, PC, explanation) into training, validation, and test sets, and use the validation set for hyperparameter tuning. Refer to the supplementary material for additional details in the experiment.

\para{Evaluation metric.} Since our problem is imbalanced,
we use the F1 score as our evaluation metric.
For the tagging approach, we average the labels of words with the same stemmed version to obtain a single prediction for the stemmed word.
To establish a baseline, we consider a random method that predicts the positive label with 0.15 probability (the base rate of positive instances).

\subsection{Prediction Performance}
\label{sec:prediction_performance}

\para{Overall performance (\figref{fig:exp_overall_perf}).}
Although our word-level task is heavily imbalanced, 
all of our models
outperform the random baseline by a wide margin.
As expected, content words are much more difficult to predict than stopwords, but the best F1 score in content words more than doubles that of the 
random baseline (0.286 vs. 0.116).
Notably, although we strongly improve on our random baseline, even our best F1 scores are relatively low, and this holds true regardless of the model used. Despite
involving
more tokens than standard tagging tasks (e.g.,
\citet{marcus1994building} and \citet{plank2016multilingual}), 
predicting whether a word is going to be echoed in explanations remains a challenging problem.
Although the vanilla LSTM model incorporates 
additional knowledge (in the form of word embeddings), the feature-based  XGBoost and logistic regression models both outperform the vanilla LSTM model.
Concatenating our proposed features with word embeddings leads to improved performance from the LSTM model, which becomes comparable to XGBoost.
This suggests that our proposed features can be difficult to learn with
an LSTM alone.

Despite the non-linearity observed in \figref{fig:copy-oppc_to_exp},
XGBoost only outperforms logistic regression by a small margin.
In the rest of this section, we use XGBoost to further examine the effectiveness of different groups of features, and model performance in different conditions.

\begin{table}[t]
\small
\centering
\begin{tabular}{l@{\hskip 1em}r@{\hskip 1em}r@{\hskip 1em}r@{\hskip 1em}r}
\toprule
& content & stop & & \\
all features & 0.286 & 0.600 & & \\
random & 0.116 & 0.205 & & \\
\midrule
& \multicolumn{2}{c}{forward} & \multicolumn{2}{c}{backward}\\
& content & stop & content & stop \\
\midrule
Non-contextual prop. & 0.177 & {\bf 0.582} & 0.285 & {\bf 0.561} \\
OP usage & 0.191  & 0.527 & 0.281 & 0.599\\
PC usage &  0.233 & 0.520 & {\bf 0.275} & 0.598 \\
OP-PC relation & {\bf 0.280} & 0.542 & 0.289 & 0.600 \\
General OP/PC prop. & 0.153 & 0.266 & 0.285 &  0.598\\
\bottomrule
\end{tabular}
\caption{Ablation performance with XGBoost on content words and stopwords (each ablated model is tuned based on performance on all words). ``forward'' refers to only using a group of features, while ``backward'' refers to only removing a group of features.
}
\label{tb:ablation}
\end{table}

\para{Ablation performance (\tableref{tb:ablation}).}
First, if we only consider a single group of features, as we hypothesized, the relation between OP and PC is crucial and leads to almost as strong performance in content words as using all features.
To further understand the strong performance of OP-PC relation, \figref{fig:exp_importance} shows the feature importance in the ablated model,
measured by the normalized total gain (see the supplementary material for feature importance in the full model).
A word's occurrence in both the OP and PC is clearly the most important feature, with distance between its POS tag distributions as the second most important.
Recall that in \tableref{tb:features} we show that words 
that have similar POS behavior between the OP and PC are more likely to be echoed in the explanation.
Overall, it seems that word-level properties contribute the most valuable signals for predicting stopwords.
If we restrict ourselves to only information in either an OP or PC, how a word is used in a PC is much more predictive of content word echoing (0.233 vs 0.191).
This observation suggests that, for content words, the PC captures more valuable information than the OP.
This finding is somewhat surprising given that the OP sets the topic of discussion and writes the explanation.

As for the effects of removing a group of features, we can see that there is little change in the performance on content words.
This can be explained by the strong performance of the OP-PC relation on its own, and the possibility of the OP-PC relation being approximated by OP and PC usage.
Again, word-level properties are valuable for strong performance in stopwords.

\para{Performance vs. word source (\figref{fig:exp_oppc_perf}).} 
We further break down the performance by where a word is from.
We can group a word based on whether it shows up only in an OP, a PC, or both OP and PC, as shown in \tableref{tb:example}.
There is a striking difference between the performance in the three categories (e.g., for all words, 0.63 in OP \& PC vs. 0.271 in PC only).
The strong performance on words in both the OP and PC applies to stopwords and content words, even accounting for the shift in the random baseline, and recalls the importance of occurring both in OP and PC as a feature.
Furthermore, the 
echoing of words from the PC is harder to predict (0.271) than from the OP (0.347) despite the fact that words only in PCs are more likely to be echoed than words only in OPs (13.5\% vs. 8.6\%). %
The performance difference is driven by stopwords, suggesting that our overall model is better at capturing signals for stopwords used in OPs.
This might relate to the fact that the OP and the explanation are written by the same author; 
prior studies 
have demonstrated the important role of stopwords for authorship attribution \citep{raghavan2010authorship}.

\para{Nouns are the most reliably predicted part-of-speech tag within content words (\tableref{tb:pos_perf}).}
Next, we break down the performance by part-of-speech tags.
We focus on the part-of-speech tags that are semantically important, namely, nouns, proper nouns, verbs, adverbs, and adjectives.

Prediction performance can be seen as a proxy for how reliably a part-of-speech tag is reused when providing explanations.
Consistent with our expectations for the importance of nouns and verbs,
our models achieve the best performance on nouns within content words.
Verbs are more challenging, but become the least difficult tag to predict when we consider all words, likely due to stopwords such as ``have.''
Adjectives turn out to be the most challenging category, suggesting that adjectival choice is perhaps more arbitrary than other parts of speech, and therefore less 
central to the process of constructing an explanation.
The important role of nouns in shaping explanations resonates with the high recall rate of nouns in 
memory tasks \citep{reynolds1976recognition}.

\begin{table}[t]
    \centering
    \small
    \begin{tabular}{p{3cm}rr|r}
        \toprule
        & content & all & random\\
        \midrule
        noun & 0.354  & 0.361 & 0.130 \\
        adverb & 0.342   & 0.411 & 0.127\\
        verb & 0.306 &  0.466  & 0.122 \\
        proper noun & 0.280 & 0.336  & 0.109 \\
        adjective & 0.237 &  0.289&  0.111\\
        \bottomrule
    \end{tabular}
    \caption{Performance on 
    five non-function part-of-speech tags (sorted by performance within content words).
    As a comparison, we also show the performance of the random baseline on content words, which is relatively stable across part-of-speech tags.
    }
    \label{tb:pos_perf}
\end{table}

\subsection{The Effect on Generating Explanations}
\label{ssec:generation}

\begin{table}[t]
\small
\begin{tabular}{lrrr}
\toprule
& ROUGE-1 & ROUGE-2 & ROUGE-L \\
\midrule
w/o features & 18.91 & 4.12 & 17.05 \\
with features & 22.01 & 3.93 & 19.02 \\
\bottomrule
\end{tabular}
\caption{
ROUGE scores (F1) on the test dataset  \citep{rouge}. The differences in ROUGE-1 and ROUGE-L are statistically significant with $p<1\mathrm{e}{-10}$.}
\label{tab:generationresults}
\end{table}

One way to measure the ultimate success of understanding pointers in explanations is to be able to generate explanations.
We use the pointer generator network with coverage  
as our starting point \cite{see_get_2017,opennmt} (see the supplementary material for details).
We investigate whether concatenating our proposed features with word embeddings can improve generation performance, as measured by ROUGE scores.

Consistent with results in sequence tagging for word-level echoing prediction, our proposed features can enhance a neural model with copying mechanisms (see \tableref{tab:generationresults}). 
Specifically, their use leads to statistically significant improvement in ROUGE-1 and ROUGE-L, while slightly hurting the performance in ROUGE-2 (the difference is not statistically significant).
We also find that our features can 
increase the likelihood of copying:
an average of 17.59 unique words get copied to the generated explanation with our features, compared to 14.17 unique words without our features. For comparison, target explanations have an average of 34.81 unique words.
We emphasize that generating explanations is a very challenging task (evidenced by the low ROUGE scores and examples in the supplementary material), and that fully solving the generation task requires more work.

\section{Concluding Discussions}
\label{sec:conclusion}

In this work, we conduct the first large-scale empirical study of everyday explanations in the context of persuasion.
We assemble a novel dataset and formulate a word-level prediction task to understand the selective nature of explanations.
Our results suggest that the relation between an OP and PC plays an important role in predicting the echoing of content words, while a word's non-contextual properties matter for stopwords.
We show that vanilla LSTMs fail to learn some of the features we develop and that our proposed features can improve the performance in generating explanations using pointer networks.
We also demonstrate the important role of nouns in shaping explanations. 

Although our approach strongly outperforms random baselines, the relatively low F1 scores indicate that predicting which word is echoed in explanations is a very challenging task.
It follows that we are only able to derive a limited understanding of how people choose to echo words in explanations. 
The extent to which explanation construction is fundamentally random \citep{nisbett1977telling}, or whether there exist other unidentified patterns, is of course an open question.
We hope that our study and the resources that we release encourage further work in understanding the pragmatics of explanations.

There are many promising research directions for future work in advancing the computational understanding of explanations.
First, 
although \cmv has the useful property that its explanations are closely connected to its explananda, 
it is important to further investigate the extent to which our findings generalize beyond \cmv and Reddit and establish universal properties of explanations.
Second, it is important to connect the words in explanations that we investigate here to the structure of explanations in pyschology \citep{lombrozo2006structure}.
Third, in addition to understanding what goes into an explanation, we need to understand what makes an explanation effective.
A better understanding of explanations not only helps develop explainable AI, but also informs the process of collecting explanations that machine learning systems learn from \citep{hancock_training_2018,rajani2019explain,camburu2018snli}.

\section*{Acknowledgments}
We thank Kimberley Buchan, anonymous reviewers, and members of the NLP+CSS research group at CU Boulder for their insightful comments and discussions; 
Jason Baumgartner for sharing the dataset that enabled this research.

\bibliography{references}

\begin{thebibliography}{54}
\expandafter\ifx\csname natexlab\endcsname\relax\def\natexlab#1{#1}\fi

\bibitem[{Blyth(1972)}]{blyth1972simpson}
Colin~R Blyth. 1972.
\newblock On {Simpson's} paradox and the sure-thing principle.
\newblock \emph{Journal of the American Statistical Association},
  67(338):364--366.

\bibitem[{Camburu et~al.(2018)Camburu, Rockt{\"a}schel, Lukasiewicz, and
  Blunsom}]{camburu2018snli}
Oana-Maria Camburu, Tim Rockt{\"a}schel, Thomas Lukasiewicz, and Phil Blunsom.
  2018.
\newblock {e-SNLI}: Natural language inference with natural language
  explanations.
\newblock In \emph{Proceedings of NeurIPS}.

\bibitem[{{Center for Language and Information
  Research}(2016)}]{clear_guidelines}
{Center for Language and Information Research}. 2016.
\newblock {ClearNLP Guidelines}.
\newblock
  \url{https://github.com/clir/clearnlp-guidelines/blob/master/md/specifications/dependency_labels.md}.
\newblock [Online; accessed 21-May-2019].

\bibitem[{Chen and Guestrin(2016)}]{chen2016xgboost}
Tianqi Chen and Carlos Guestrin. 2016.
\newblock {XGBoost}: A scalable tree boosting system.
\newblock In \emph{Proceedings of KDD}.

\bibitem[{{CMV moderators}(2019{\natexlab{a}})}]{cmvmedia}
{CMV moderators}. 2019{\natexlab{a}}.
\newblock {CMV media coverage}.
\newblock \url{https://changemyview.net/subreddit/#media-coverage}.
\newblock [Online; accessed 27-Apr-2019].

\bibitem[{{CMV moderators}(2019{\natexlab{b}})}]{deltasystem}
{CMV moderators}. 2019{\natexlab{b}}.
\newblock {The Delta System}.
\newblock \url{https://www.reddit.com/r/changemyview/wiki/deltasystem}.
\newblock [Online; accessed 27-Apr-2019].

\bibitem[{Danescu-Niculescu-Mizil et~al.(2011)Danescu-Niculescu-Mizil, Gamon,
  and Dumais}]{Danescu-Niculescu-Mizil:2011:MMW:1963405.1963509}
Cristian Danescu-Niculescu-Mizil, Michael Gamon, and Susan Dumais. 2011.
\newblock Mark my words!: Linguistic style accommodation in social media.
\newblock In \emph{Proceedings of WWW}.

\bibitem[{Danescu-Niculescu-Mizil et~al.()Danescu-Niculescu-Mizil, Lee, Pang,
  and Kleinberg}]{echoes}
Cristian Danescu-Niculescu-Mizil, Lillian Lee, Bo~Pang, and Jon Kleinberg.
\newblock Echoes of power: Language effects and power differences in social
  interaction.
\newblock In \emph{Proceedings of WWW}.

\bibitem[{Doshi-Velez and Kim(2017)}]{doshi2017towards}
Finale Doshi-Velez and Been Kim. 2017.
\newblock Towards a rigorous science of interpretable machine learning.
\newblock \emph{arXiv preprint arXiv:1702.08608}.

\bibitem[{Durmus and Cardie(2018)}]{esin}
Esin Durmus and Claire Cardie. 2018.
\newblock Exploring the role of prior beliefs for argument persuasion.
\newblock In \emph{Proceedings of NAACL}.

\bibitem[{Giles and Ogay(2007)}]{giles2007communication}
Howard Giles and Tania Ogay. 2007.
\newblock Communication accommodation theory.
\newblock \emph{Explaining communication: Contemporary theories and exemplars},
  pages 293--310.

\bibitem[{Gu et~al.(2016)Gu, Lu, Li, and Li}]{gu_incorporating_2016}
Jiatao Gu, Zhengdong Lu, Hang Li, and Victor O.~K. Li. 2016.
\newblock \href {https://www.aclweb.org/anthology/P16-1154} {Incorporating
  {Copying} {Mechanism} in {Sequence}-to-{Sequence} {Learning}}.
\newblock In \emph{Proceedings of ACL}.

\bibitem[{Guidotti et~al.(2019)Guidotti, Monreale, Ruggieri, Turini, Giannotti,
  and Pedreschi}]{guidotti2019survey}
Riccardo Guidotti, Anna Monreale, Salvatore Ruggieri, Franco Turini, Fosca
  Giannotti, and Dino Pedreschi. 2019.
\newblock A survey of methods for explaining black box models.
\newblock \emph{ACM computing surveys (CSUR)}, 51(5):93.

\bibitem[{Habernal and Gurevych(2016)}]{habernal2016makes}
Ivan Habernal and Iryna Gurevych. 2016.
\newblock What makes a convincing argument? {Empirical} analysis and detecting
  attributes of convincingness in web argumentation.
\newblock In \emph{Proceedings of EMNLP}.

\bibitem[{Hancock et~al.(2018)Hancock, Varma, Wang, Bringmann, Liang, and
  Ré}]{hancock_training_2018}
Braden Hancock, Paroma Varma, Stephanie Wang, Martin Bringmann, Percy Liang,
  and Christopher Ré. 2018.
\newblock \href {https://arxiv.org/abs/1805.03818} {Training {Classifiers} with
  {Natural} {Language} {Explanations}}.
\newblock In \emph{Proceedings of ACL}.

\bibitem[{Heffernan(2018)}]{heffernan:2018a}
Virgina Heffernan. 2018.
\newblock \href
  {https://www.wired.com/story/free-speech-issue-reddit-change-my-view/} {Our
  best hope for civil discourse online is on ... {Reddit}}.
\newblock \emph{Wired}.

\bibitem[{Hempel and Oppenheim(1948)}]{hempel1948studies}
Carl~G Hempel and Paul Oppenheim. 1948.
\newblock Studies in the logic of explanation.
\newblock \emph{Philosophy of science}, 15(2):135--175.

\bibitem[{Honnibal and Montani(2017)}]{honnibal_spacy_2017}
Matthew Honnibal and Ines Montani. 2017.
\newblock {spaCy} 2: {Natural} language understanding with {Bloom} embeddings,
  {convolutional} {neural} {networks} and incremental parsing.

\bibitem[{Horne et~al.(2019)Horne, Nevo, O'Donovan, Cho, and
  Adali}]{horne2019rating}
Benjamin~D Horne, Dorit Nevo, John O'Donovan, Jin-Hee Cho, and Sibel Adali.
  2019.
\newblock Rating reliability and bias in news articles: Does ai assistance help
  everyone?
\newblock In \emph{Proceedings of ICWSM}.

\bibitem[{Keil(2006)}]{keil2006explanation}
Frank~C Keil. 2006.
\newblock Explanation and understanding.
\newblock \emph{Annu. Rev. Psychol.}, 57:227--254.

\bibitem[{Kingma and Ba(2015)}]{kingma2015adam}
Diederik~P. Kingma and Jimmy Ba. 2015.
\newblock Adam: A method for stochastic optimization.
\newblock In \emph{Proceedings of ICLR}.

\bibitem[{Klein et~al.(2017)Klein, Kim, Deng, Senellart, and Rush}]{opennmt}
Guillaume Klein, Yoon Kim, Yuntian Deng, Jean Senellart, and Alexander~M. Rush.
  2017.
\newblock \href {https://doi.org/10.18653/v1/P17-4012} {Open{NMT}: Open-source
  toolkit for neural machine translation}.
\newblock In \emph{Proceedings of ACL}.

\bibitem[{Lai and Tan(2019)}]{lai+tan:19}
Vivian Lai and Chenhao Tan. 2019.
\newblock On human predictions with explanations and predictions of machine
  learning models: A case study on deception detection.
\newblock In \emph{Proceedings of FAT*}.

\bibitem[{Leskovec et~al.(2009)Leskovec, Backstrom, and
  Kleinberg}]{leskovec2009meme}
Jure Leskovec, Lars Backstrom, and Jon Kleinberg. 2009.
\newblock Meme-tracking and the dynamics of the news cycle.
\newblock In \emph{Proceedings of KDD}.

\bibitem[{Lin(2004)}]{rouge}
Chin-Yew Lin. 2004.
\newblock \href {https://www.aclweb.org/anthology/W04-1013} {{ROUGE}: A package
  for automatic evaluation of summaries}.
\newblock In \emph{Text Summarization Branches Out: Proceedings of the {ACL}-04
  Workshop}, pages 74--81, Barcelona, Spain. Association for Computational
  Linguistics.

\bibitem[{Ling et~al.(2016)Ling, Blunsom, Grefenstette, Hermann,
  Ko{\v{c}}isk{\'y}, Wang, and Senior}]{ling-etal-2016-latent}
Wang Ling, Phil Blunsom, Edward Grefenstette, Karl~Moritz Hermann,
  Tom{\'a}{\v{s}} Ko{\v{c}}isk{\'y}, Fumin Wang, and Andrew Senior. 2016.
\newblock \href {https://doi.org/10.18653/v1/P16-1057} {Latent predictor
  networks for code generation}.
\newblock In \emph{Proceedings of ACL}, pages 599--609, Berlin, Germany.
  Association for Computational Linguistics.

\bibitem[{Lippi and Torroni(2016)}]{lippi2016argumentation}
Marco Lippi and Paolo Torroni. 2016.
\newblock Argumentation mining: State of the art and emerging trends.
\newblock \emph{ACM Transactions on Internet Technology (TOIT)}, 16(2):10.

\bibitem[{Lipton(2016)}]{lipton2016mythos}
Zachary~C Lipton. 2016.
\newblock The mythos of model interpretability.
\newblock \emph{arXiv preprint arXiv:1606.03490}.

\bibitem[{Lombrozo(2006)}]{lombrozo2006structure}
Tania Lombrozo. 2006.
\newblock The structure and function of explanations.
\newblock \emph{Trends in cognitive sciences}, 10(10):464--470.

\bibitem[{Loper and Bird(2002)}]{loper2002nltk}
Edward Loper and Steven Bird. 2002.
\newblock {NLTK}: the natural language toolkit.
\newblock \emph{arXiv preprint cs/0205028}.

\bibitem[{Lukin et~al.(2017)Lukin, Anand, Walker, and
  Whittaker}]{lukin2017argument}
Stephanie~M Lukin, Pranav Anand, Marilyn Walker, and Steve Whittaker. 2017.
\newblock Argument strength is in the eye of the beholder: Audience effects in
  persuasion.
\newblock In \emph{Proceedings of EACL}.

\bibitem[{Marcus et~al.(1994)Marcus, Santorini, and
  Marcinkiewicz}]{marcus1994building}
Mitchell~P. Marcus, Beatrice Santorini, and Mary~A. Marcinkiewicz. 1994.
\newblock Building a large annotated corpus of english: The penn treebank.
\newblock \emph{Computational Linguistics}, 19:313--330.

\bibitem[{Miller(2018)}]{miller2018explanation}
Tim Miller. 2018.
\newblock Explanation in artificial intelligence: Insights from the social
  sciences.
\newblock \emph{Artificial Intelligence}.

\bibitem[{Nisbett and Wilson(1977)}]{nisbett1977telling}
Richard~E Nisbett and Timothy~D Wilson. 1977.
\newblock Telling more than we can know: Verbal reports on mental processes.
\newblock \emph{Psychological review}, 84(3):231.

\bibitem[{Pennington et~al.(2014)Pennington, Socher, and
  Manning}]{pennington_glove:_2014}
Jeffrey Pennington, Richard Socher, and Christopher Manning. 2014.
\newblock \href {https://doi.org/10.3115/v1/D14-1162} {{GloVe}: {Global}
  {Vectors} for {Word} {Representation}}.
\newblock In \emph{Proceedings of EMNLP}, pages 1532--1543, Doha, Qatar.
  Association for Computational Linguistics.

\bibitem[{Plank et~al.(2016)Plank, S{\o}gaard, and
  Yoav}]{plank2016multilingual}
Barbara Plank, Anders S{\o}gaard, and Goldberg Yoav. 2016.
\newblock Multilingual part-of-speech tagging with bidirectional long
  short-term memory models and auxiliary loss.
\newblock In \emph{Proceedings of ACL (short papers)}.

\bibitem[{Porter(1980)}]{porter1980algorithm}
Martin~F. Porter. 1980.
\newblock An algorithm for suffix stripping.
\newblock \emph{Program}, 14(2):130--137.

\bibitem[{Raghavan et~al.(2010)Raghavan, Kovashka, and
  Mooney}]{raghavan2010authorship}
Sindhu Raghavan, Adriana Kovashka, and Raymond Mooney. 2010.
\newblock Authorship attribution using probabilistic context-free grammars.
\newblock In \emph{Proceedings of ACL (short papers)}, pages 38--42.

\bibitem[{Rajani et~al.(2019)Rajani, McCann, Xiong, and
  Socher}]{rajani2019explain}
Nazneen~Fatema Rajani, Bryan McCann, Caiming Xiong, and Richard Socher. 2019.
\newblock Explain yourself! leveraging language models for commonsense
  reasoning.
\newblock In \emph{Proceedings of ACL}.

\bibitem[{Reynolds and Flagg(1976)}]{reynolds1976recognition}
Allan~G Reynolds and Paul~W Flagg. 1976.
\newblock Recognition memory for elements of sentences.
\newblock \emph{Memory \& Cognition}, 4(4):422--432.

\bibitem[{Ribeiro et~al.(2016)Ribeiro, Singh, and Guestrin}]{ribeiro2016should}
Marco~Tulio Ribeiro, Sameer Singh, and Carlos Guestrin. 2016.
\newblock Why should {I} trust you?: Explaining the predictions of any
  classifier.
\newblock In \emph{Proceedings of KDD}.

\bibitem[{Salmon(2006)}]{salmon2006four}
Wesley~C Salmon. 2006.
\newblock \emph{Four decades of scientific explanation}.
\newblock University of Pittsburgh press.

\bibitem[{Schuster and Manning(2016)}]{schuster_enhanced_2016}
Sebastian Schuster and Christopher Manning. 2016.
\newblock \href
  {http://nlp.stanford.edu/~sebschu/pubs/schuster-manning-lrec2016.pdf}
  {Enhanced {English} {Universal} {Dependencies}: an {Improved}
  {Representation} for {Natural} {Language} {Understanding} {Tasks}}.
\newblock \emph{LREC 2016}.

\bibitem[{See et~al.(2017)See, Liu, and Manning}]{see_get_2017}
Abigail See, Peter~J. Liu, and Christopher~D. Manning. 2017.
\newblock \href {http://arxiv.org/abs/1704.04368} {Get {To} {The} {Point}:
  {Summarization} with {Pointer}-{Generator} {Networks}}.
\newblock In \emph{Proceedings of ACL}.

\bibitem[{Simmons et~al.(2011)Simmons, Adamic, and Adar}]{simmons2011memes}
Matthew~P Simmons, Lada~A Adamic, and Eytan Adar. 2011.
\newblock Memes online: Extracted, subtracted, injected, and recollected.
\newblock In \emph{Proceedings of ICWSM}.

\bibitem[{Somasundaran and Wiebe(2009)}]{somasundaran2009recognizing}
Swapna Somasundaran and Janyce Wiebe. 2009.
\newblock Recognizing stances in online debates.
\newblock In \emph{Proceedings of ACL}.

\bibitem[{Tan et~al.(2016)Tan, Niculae, Danescu-Niculescu-Mizil, and
  Lee}]{tan_winning_2016}
Chenhao Tan, Vlad Niculae, Cristian Danescu-Niculescu-Mizil, and Lillian Lee.
  2016.
\newblock \href {https://doi.org/10.1145/2872427.2883081} {Winning {Arguments}:
  {Interaction} {Dynamics} and {Persuasion} {Strategies} in {Good}-faith
  {Online} {Discussions}}.
\newblock In \emph{Proceedings of WWW}, {WWW} '16, pages 613--624, Republic and
  Canton of Geneva, Switzerland. International World Wide Web Conferences
  Steering Committee.
\newblock Event-place: Montréal, Québec, Canada.

\bibitem[{Tan et~al.(2018)Tan, Peng, and Smith}]{chenhaopres}
Chenhao Tan, Hao Peng, and Noah~A. Smith. 2018.
\newblock You are no {Jack Kennedy}: On media selection of highlights from
  presidential debates.
\newblock In \emph{Proceedings of WWW}.

\bibitem[{Vinyals et~al.(2015)Vinyals, Fortunato, and
  Jaitly}]{vinyals2015pointer}
Oriol Vinyals, Meire Fortunato, and Navdeep Jaitly. 2015.
\newblock Pointer networks.
\newblock In \emph{Proceedings of NeurIPS}, pages 2692--2700.

\bibitem[{Walker et~al.(2012)Walker, Anand, Abbott, and
  Grant}]{walker2012stance}
Marilyn~A Walker, Pranav Anand, Robert Abbott, and Ricky Grant. 2012.
\newblock Stance classification using dialogic properties of persuasion.
\newblock In \emph{Proceedings of NAACL}.

\bibitem[{Wang et~al.(2017)Wang, Beauchamp, Shugars, and Qin}]{wang2017winning}
Lu~Wang, Nick Beauchamp, Sarah Shugars, and Kechen Qin. 2017.
\newblock Winning on the merits: The joint effects of content and style on
  debate outcomes.
\newblock \emph{Transactions of the Association for Computational Linguistics}.

\bibitem[{Wilson and Keil(1998)}]{wilson_shadows_1998}
Robert~A. Wilson and Frank Keil. 1998.
\newblock \href {https://doi.org/10.1023/A:1008259020140} {The {Shadows} and
  {Shallows} of {Explanation}}.
\newblock \emph{Minds and Machines}, 8(1):137--159.

\bibitem[{Woodward(2005)}]{woodward2005making}
James Woodward. 2005.
\newblock \emph{Making things happen: A theory of causal explanation}.
\newblock Oxford university press.

\bibitem[{Zhang et~al.(2016)Zhang, Kumar, Ravi, and
  Danescu-Niculescu-Mizil}]{justine}
Justine Zhang, Ravi Kumar, Sujith Ravi, and Cristian Danescu-Niculescu-Mizil.
  2016.
\newblock Conversational flow in {Oxford}-style debates.
\newblock In \emph{Proceedings of NAACL}.

\end{thebibliography}
\bibliographystyle{acl_natbib}

\newpage

\appendix

\section{Supplemental Material}

\subsection{Preprocessing.} Before tokenizing, we pass each OP, PC, and explanation through a preprocessing pipeline, with the following steps:
\begin{enumerate}
    \item Occasionally, \cmv's moderators will edit comments, prefixing their edits with ``Hello, users of CMV'' or ``This is a footnote'' (see \tableref{tab:sample_data}). We remove this, and any text that follows on the same line.
    \item We replace URLs with a ``@url@'' token, defining a URL to be any string which matches the following regular expression: \verb|(https?://[^\s)]*)|.
    \item We replace ``$\Delta$'' symbols and their analogues---such as ``$\delta$'', ``\&;\#8710;'', and ``!delta''---with the word ``delta''. We also remove the word ``delta'' from explanations, if the explanation starts with delta.
    \item Reddit--specific prefixes, such as ``u/'' (denoting a user) and ``r/'' (denoting a subreddit) are removed, as we observed that they often interfered with spaCy's ability to correctly parse its inputs.
    \item We remove any text matching the regular expression \verb|EDIT(.*?):.*| from the beginning of the match to the end of that line, as well as variations, such as \verb|Edit(.*?):.*|.
    \item Reddit allows users to insert blockquoted text. We extract any blockquotes and surround them with standard quotation marks.
    \item We replace all contiguous whitespace with a single space. We also do this with tab characters and carriage returns, and with two or more hyphens, asterisks, or underscores.
\end{enumerate}

\begin{table}[t]
\centering
\small
\begin{tabular}{p{0.9\columnwidth}}
\toprule
\textbf{Sample footnote}: Hello, users of CMV! This is a footnote from your moderators. We’d just like to remind you of a couple of things. Firstly, please remember to read through our rules. If you see a comment that has broken one, it is more effective to report it than downvote it. Speaking of which, *downvotes don’t change views**! If you are thinking about submitting a CMV yourself, please have a look through our **popular topics wiki first. Any questions or concerns? Feel free to message us**. Happy CMVing!*\\
\textbf{Sample subreddit reference}: r/ideasforcmv, /r/nba\\
\textbf{Sample URL} : https://www.quora.com/profile/\\
\textbf{Sample user reference}: u/Ansuz07\\
\textbf{Sample edit}: EDIT for clarification: This isn't to suggest that you have to remain financially independent to vote\\
\bottomrule
\end{tabular}

\caption{Sample data that were affected by preprocessing.}
\label{tab:sample_data}
\end{table}

\para{Tokenizing the data.}  After passing text through our preprocessing pipeline, 
we use the default spaCy 
pipeline 
to extract part-of-speech tags, dependency tags, and entity details for each token\footnote{We ignore all tokens tagged as "SPACE" by the part of speech tagger.} \citep{honnibal_spacy_2017}. In addition, we use NLTK to stem words \citep{loper2002nltk}. This is used to compute all word level features discussed in Section 4 of the main paper.

\subsection{PC Echoing OP}

\begin{figure*}[t]
\begin{subfigure}[t]{0.32\textwidth}
  \includegraphics[width=\textwidth]{figures/png/copy-oppc_to_exp.png}
  \caption{Echoing probability between explananda and explanations.}
  \label{fig:copy-oppc_to_exp}
\end{subfigure}
\begin{subfigure}[t]{0.32\textwidth}
  \includegraphics[width=\textwidth]{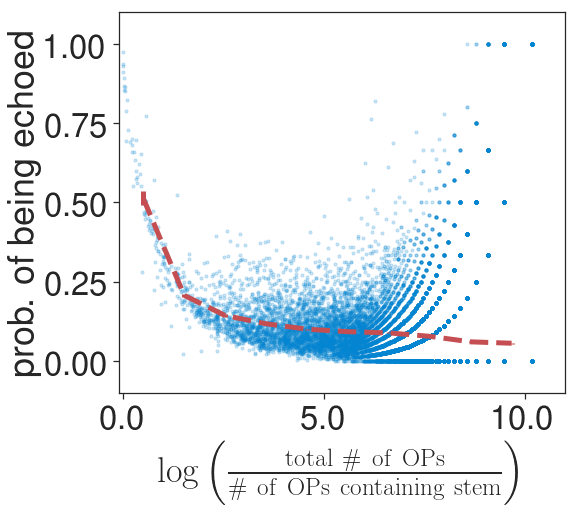}
  \caption{Echoing probability between OPs and their PCs.}
  \label{fig:copy-op_to_pc}
\end{subfigure}
\begin{subfigure}[t]{0.32\textwidth}
  \includegraphics[width=\textwidth]{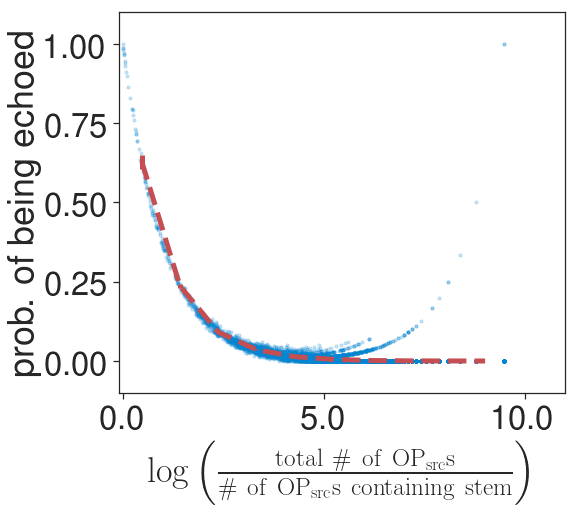}
  \caption{Echoing probability between OPs and other, randomly chosen OPs.}
  \label{fig:copy-op_to_op}
\end{subfigure}
\caption{The U-shape exists both in \figref{fig:copy-oppc_to_exp} and \figref{fig:copy-op_to_pc}, but not in \figref{fig:copy-op_to_op}.}
\end{figure*}

\figref{fig:copy-op_to_pc} shows a similar U-shape in the probability of a word being echoed in PC.
However, visually, we can see that rare words seem more likely to have high echoing probability in explanations, while that probability is higher for words with moderate frequency in PCs.
As PCs tend to be longer than explanations, we also used the echoing probability of the most frequent words to normalize the probability of other words so that they are comparable.
We indeed observed a higher likelihood of echoing the rare words, but lower likelihood of echoing words with moderate frequency in explanations than in PCs.

\subsection{Feature Calculation}

Given an OP, PC, and explanation, we calculate a 66--dimensional vector for each unique stem in the concatenated OP and PC. Here, we describe the process of calculating each feature.

\begin{enumerate}[leftmargin=0.5in]
\item[1.] Inverse document frequency: for a stem $s$, the inverse document frequency is given by $\log \frac{N}{\mathrm{df}_s}$, where $N$ is the total number of documents (here, OPs and PCs) in the training set, and $\mathrm{df}_s$ is the number of documents in the training data whose set of stemmed words contains $s$.
\item[2.] Stem length: the number of characters in the stem.
\item[3.] Wordnet depth (min): starting with the stem, this is the length of the minimum hypernym path to the synset root.
\item[4.] Wordnet depth (max): similarly, this is the length of the maximum hypernym path.
\item[5.] Stem transfer probability: the percentage of times in which a stem seen in the explanandum is also seen in the explanation. If, during validation or testing, a stem is encountered for the first time, we set this to be the mean probability of transfer over all stems seen in the training data.
\item[6-21.] OP part--of--speech tags: a stem can represent multiple parts of speech. For example, both ``traditions'' and ``traditional'' will be stemmed to ``tradit.'' We count the percentage of times the given stem appears as each part--of--speech tag, following the Universal Dependencies scheme \citep{schuster_enhanced_2016}.\footnote{Note that, for English, spaCy does not use the \texttt{SCONJ} tag.} If the stem does not appear in the OP, each part--of--speech feature will be $\frac{1}{16}$.
\item[22-24.] OP subject, object, and other: Given a stem $s$, we calculate the percentage of times that $s$'s surface forms in the OP are classified as subjects, objects, or something else by SpaCy. We follow the CLEAR guidelines, \citep{clear_guidelines} and use the following tags to indicate a subject: \texttt{nsubj}, \texttt{nsubjpass}, \texttt{csubj}, \texttt{csubjpass}, \texttt{agent}, and \texttt{expl}. Objects are identified using these tags: \texttt{dobj}, \texttt{dative}, \texttt{attr}, \texttt{oprd}. If $s$ does not appear at all in the OP, we let subject, object, and other each equal $\frac{1}{3}$.
\item[25.] OP term frequency: the number of times any surface form of a stem appears in the list of tokens that make up the OP.
\item[26.] OP normalized term frequency: the percentage of the OP's tokens which are a surface form of the given stem.
\item[27.] OP \# of surface forms: the number of different surface forms for the given stem.
\item[28.] OP location: the average location of each surface form of the given stem which appears in the OP, where the location of a surface form is defined as the percentage of tokens which appear after that surface form. If the stem does not appear at all in the OP, this value is $\frac{1}{2}$. 
\item[29.] OP is in quotes: the number of times the stem appears in the OP surrounded by quotation marks.
\item[30.] OP is entity: the percentage of tokens in the OP that are both a surface form for the given stem, and are tagged by SpaCy as one of the following entities: \texttt{PERSON}, \texttt{NORP}, \texttt{FAC}, \texttt{ORG}, \texttt{GPE}, \texttt{LOC}, \texttt{PRODUCT}, \texttt{EVENT}, \texttt{WORK\_OF\_ART}, \texttt{LAW}, and \texttt{LANGUAGE}.
\item[31-55.] PC equivalents of features 6-30.
\item[56.] In both OP and PC: 1, if one of the stem's surface forms appears in both the OP and PC. 0 otherwise.
\item[57.] \# of unique surface forms in OP: for the given stem, the number of surface forms that appear in the OP, but not in the PC.
\item[58.] \# of unique surface forms in PC: for the given stem, the number of surface forms that appear in the PC, but not in the OP.
\item[59.] Stem part--of--speech distribution difference: we consider the concatenation of features 6-21, along with the concatenation of features 31-46, as two distributions, and calculate the Jensen--Shannon divergence between them.
\item[60.] Stem dependency distribution difference: similarly, we consider the concatenation of features 22-24 (OP dependency labels), and the concatenation of features 47-49 (PC dependency labels), as two distributions, and calculate the Jensen--Shannon divergence between them.
\item[61.] OP length: the number of tokens in the OP.
\item[62.] PC length: the number of tokens in the PC.
\item[63.] Length difference: the absolute value of the difference between OP length and PC length.
\item[64.] Avg.\ word length difference: the difference between the average number of characters per token in the OP and the average number of characters per token in the PC.
\item[65.] OP/PC part--of--speech tag distribution difference: the Jensen--Shannon divergence between the part--of--speech tag distributions of the OP on the one hand, and the PC on the other.
\item[66.] Depth of the PC in the thread: since there can be many back--and--forth replies before a user awards a delta, we number each comment in a thread, starting at 0 for the OP, and incrementing for each new comment before the PC appears.
\end{enumerate}

\begin{table*}[t]
\centering
\small

\begin{tabularx}{0.9\textwidth}{Xrrr}
\toprule
Feature & all & content words & stopwords\\
\midrule
Inverse document frequency & $\downarrow\downarrow\downarrow\downarrow$ & $\downarrow\downarrow\downarrow\downarrow$ & $\downarrow\downarrow\downarrow\downarrow$\\
Stem length & $\downarrow\downarrow\downarrow\downarrow$ & $\downarrow\downarrow\downarrow\downarrow$ & $\downarrow\downarrow\downarrow\downarrow$\\
Wordnet depth (min) & $\downarrow\downarrow\downarrow\downarrow$ & $\uparrow\uparrow\uparrow\uparrow$ & $\uparrow\uparrow\uparrow\uparrow$\\
Wordnet depth (max) & $\downarrow\downarrow\downarrow\downarrow$ & $\uparrow\uparrow\uparrow\uparrow$ & $\uparrow\uparrow\uparrow\uparrow$\\
Stem transfer probability & $\uparrow\uparrow\uparrow\uparrow$ & $\uparrow\uparrow\uparrow\uparrow$ & $\uparrow\uparrow\uparrow\uparrow$\\
\midrule
OP ADP & $\uparrow\uparrow\uparrow\uparrow$ & $\uparrow\uparrow\uparrow\uparrow$ & $\uparrow\uparrow\uparrow\uparrow$\\
OP PRON & $\uparrow\uparrow\uparrow\uparrow$ & $\uparrow\uparrow\uparrow\uparrow$ & $\uparrow\uparrow\uparrow\uparrow$\\
OP X & $\downarrow\downarrow\downarrow\downarrow$ & $\uparrow\uparrow\uparrow\uparrow$ & $\downarrow\downarrow\downarrow\downarrow$\\
OP DET & $\uparrow\uparrow\uparrow\uparrow$ & $\uparrow\uparrow\uparrow\uparrow$ & $\uparrow\uparrow\uparrow\uparrow$\\
OP ADJ & $\downarrow\downarrow\downarrow\downarrow$ & $\downarrow\downarrow\downarrow\downarrow$ & $\downarrow\downarrow\downarrow\downarrow$\\
OP PROPN & $\downarrow\downarrow\downarrow\downarrow$ & $\downarrow\downarrow\downarrow\downarrow$ & $\downarrow\downarrow\downarrow\downarrow$\\
OP VERB & $\downarrow\downarrow\downarrow\downarrow$ & $\textcolor{white}{\downarrow}\textcolor{white}{\downarrow}\downarrow\downarrow$ & $\uparrow\uparrow\uparrow\uparrow$\\
OP PART & $\uparrow\uparrow\uparrow\uparrow$ & $\uparrow\uparrow\uparrow\uparrow$ & $\uparrow\uparrow\uparrow\uparrow$\\
OP CCONJ & $\uparrow\uparrow\uparrow\uparrow$ & $\uparrow\uparrow\uparrow\uparrow$ & $\uparrow\uparrow\uparrow\uparrow$\\
OP INTJ & $\downarrow\downarrow\downarrow\downarrow$ & $\uparrow\uparrow\uparrow\uparrow$ & $\downarrow\downarrow\downarrow\downarrow$\\
OP NOUN & $\downarrow\downarrow\downarrow\downarrow$ & $\downarrow\downarrow\downarrow\downarrow$ & $\downarrow\downarrow\downarrow\downarrow$\\
OP NUM & $\downarrow\downarrow\downarrow\downarrow$ & $\textcolor{white}{\downarrow}\textcolor{white}{\downarrow}\downarrow\downarrow$ & $\downarrow\downarrow\downarrow\downarrow$\\
OP ADV & $\downarrow\downarrow\downarrow\downarrow$ & $\uparrow\uparrow\uparrow\uparrow$ & $\downarrow\downarrow\downarrow\downarrow$\\
OP PUNCT & $\downarrow\downarrow\downarrow\downarrow$ & $\uparrow\uparrow\uparrow\uparrow$ & $\downarrow\downarrow\downarrow\downarrow$\\
OP SYM & $\downarrow\downarrow\downarrow\downarrow$ & $\uparrow\uparrow\uparrow\uparrow$ & $\downarrow\downarrow\downarrow\downarrow$\\
OP AUX & $\downarrow\downarrow\downarrow\downarrow$ & $\uparrow\uparrow\uparrow\uparrow$ & $\downarrow\downarrow\downarrow\downarrow$\\
OP subject & $\uparrow\uparrow\uparrow\uparrow$ & $\uparrow\uparrow\uparrow\uparrow$ & $\uparrow\uparrow\uparrow\uparrow$\\
OP object & $\downarrow\downarrow\downarrow\downarrow$ & $\uparrow\uparrow\uparrow\uparrow$ & $\downarrow\downarrow\downarrow\downarrow$\\
OP other & $\uparrow\uparrow\uparrow\uparrow$ & $\downarrow\downarrow\downarrow\downarrow$ & $\uparrow\uparrow\uparrow\uparrow$\\
OP term frequency & $\uparrow\uparrow\uparrow\uparrow$ & $\uparrow\uparrow\uparrow\uparrow$ & $\uparrow\uparrow\uparrow\uparrow$\\
OP normalized term frequency & $\uparrow\uparrow\uparrow\uparrow$ & $\uparrow\uparrow\uparrow\uparrow$ & $\uparrow\uparrow\uparrow\uparrow$\\
OP \# of surface forms & $\uparrow\uparrow\uparrow\uparrow$ & $\uparrow\uparrow\uparrow\uparrow$ & $\uparrow\uparrow\uparrow\uparrow$\\
OP location & $\uparrow\uparrow\uparrow\uparrow$ & $\uparrow\uparrow\uparrow\uparrow$ & ------\\
OP in quotes & $\uparrow\uparrow\uparrow\uparrow$ & $\uparrow\uparrow\uparrow\uparrow$ & $\uparrow\uparrow\uparrow\uparrow$\\
OP is entity & $\uparrow\uparrow\uparrow\uparrow$ & $\uparrow\uparrow\uparrow\uparrow$ & $\uparrow\uparrow\uparrow\uparrow$\\
\midrule
PC ADP & $\uparrow\uparrow\uparrow\uparrow$ & $\downarrow\downarrow\downarrow\downarrow$ & $\uparrow\uparrow\uparrow\uparrow$\\
PC PRON & $\uparrow\uparrow\uparrow\uparrow$ & $\downarrow\downarrow\downarrow\downarrow$ & $\uparrow\uparrow\uparrow\uparrow$\\
PC X & $\downarrow\downarrow\downarrow\downarrow$ & $\downarrow\downarrow\downarrow\downarrow$ & $\downarrow\downarrow\downarrow\downarrow$\\
PC DET & $\uparrow\uparrow\uparrow\uparrow$ & $\downarrow\downarrow\downarrow\downarrow$ & $\uparrow\uparrow\uparrow\uparrow$\\
PC ADJ & $\downarrow\downarrow\downarrow\downarrow$ & $\uparrow\uparrow\uparrow\uparrow$ & $\downarrow\downarrow\downarrow\downarrow$\\
PC PROPN & $\downarrow\downarrow\downarrow\downarrow$ & $\downarrow\downarrow\downarrow\downarrow$ & $\downarrow\downarrow\downarrow\downarrow$\\
PC VERB & $\uparrow\uparrow\uparrow\uparrow$ & $\uparrow\uparrow\uparrow\uparrow$ & $\uparrow\uparrow\uparrow\uparrow$\\
PC PART & $\downarrow\downarrow\downarrow\downarrow$ & $\downarrow\downarrow\downarrow\downarrow$ & $\uparrow\uparrow\uparrow\uparrow$\\
PC CCONJ & $\uparrow\uparrow\uparrow\uparrow$ & $\downarrow\downarrow\downarrow\downarrow$ & $\uparrow\uparrow\uparrow\uparrow$\\
PC INTJ & $\downarrow\downarrow\downarrow\downarrow$ & $\downarrow\downarrow\downarrow\downarrow$ & $\downarrow\downarrow\downarrow\downarrow$\\
PC NOUN & $\downarrow\downarrow\downarrow\downarrow$ & $\uparrow\uparrow\uparrow\uparrow$ & $\downarrow\downarrow\downarrow\downarrow$\\
PC NUM & $\downarrow\downarrow\downarrow\downarrow$ & $\downarrow\downarrow\downarrow\downarrow$ & $\downarrow\downarrow\downarrow\downarrow$\\
PC ADV & ------ & $\uparrow\uparrow\uparrow\uparrow$ & $\downarrow\downarrow\downarrow\downarrow$\\
PC PUNCT & $\downarrow\downarrow\downarrow\downarrow$ & $\downarrow\downarrow\downarrow\downarrow$ & $\downarrow\downarrow\downarrow\downarrow$\\
PC SYM & $\downarrow\downarrow\downarrow\downarrow$ & $\downarrow\downarrow\downarrow\downarrow$ & $\downarrow\downarrow\downarrow\downarrow$\\
PC AUX & $\downarrow\downarrow\downarrow\downarrow$ & $\downarrow\downarrow\downarrow\downarrow$ & $\downarrow\downarrow\downarrow\downarrow$\\
PC subject & $\downarrow\downarrow\downarrow\downarrow$ & $\downarrow\downarrow\downarrow\downarrow$ & $\downarrow\downarrow\downarrow\downarrow$\\
PC object & $\downarrow\downarrow\downarrow\downarrow$ & $\downarrow\downarrow\downarrow\downarrow$ & $\downarrow\downarrow\downarrow\downarrow$\\
PC other & $\uparrow\uparrow\uparrow\uparrow$ & $\uparrow\uparrow\uparrow\uparrow$ & $\uparrow\uparrow\uparrow\uparrow$\\
PC term frequency & $\uparrow\uparrow\uparrow\uparrow$ & $\uparrow\uparrow\uparrow\uparrow$ & $\uparrow\uparrow\uparrow\uparrow$\\
PC normalized term frequency & $\uparrow\uparrow\uparrow\uparrow$ & $\uparrow\uparrow\uparrow\uparrow$ & $\uparrow\uparrow\uparrow\uparrow$\\
PC \# of surface forms & $\uparrow\uparrow\uparrow\uparrow$ & $\uparrow\uparrow\uparrow\uparrow$ & $\uparrow\uparrow\uparrow\uparrow$\\
PC location & $\textcolor{white}{\uparrow}\textcolor{white}{\uparrow}\textcolor{white}{\uparrow}\uparrow$ & $\uparrow\uparrow\uparrow\uparrow$ & $\downarrow\downarrow\downarrow\downarrow$\\
PC in quotes & $\uparrow\uparrow\uparrow\uparrow$ & $\uparrow\uparrow\uparrow\uparrow$ & $\uparrow\uparrow\uparrow\uparrow$\\
PC is entity & $\uparrow\uparrow\uparrow\uparrow$ & $\uparrow\uparrow\uparrow\uparrow$ & $\uparrow\uparrow\uparrow\uparrow$\\
\midrule
In both OP and PC & $\uparrow\uparrow\uparrow\uparrow$ & $\uparrow\uparrow\uparrow\uparrow$ & $\uparrow\uparrow\uparrow\uparrow$\\
\# of unique surface forms in OP & $\downarrow\downarrow\downarrow\downarrow$ & $\downarrow\downarrow\downarrow\downarrow$ & $\downarrow\downarrow\downarrow\downarrow$\\
\# of unique surface forms in PC & $\uparrow\uparrow\uparrow\uparrow$ & $\uparrow\uparrow\uparrow\uparrow$ & $\downarrow\downarrow\downarrow\downarrow$\\
Stem POS distribution difference & $\downarrow\downarrow\downarrow\downarrow$ & $\downarrow\downarrow\downarrow\downarrow$ & $\downarrow\downarrow\downarrow\downarrow$\\
Stem dependency distribution difference & $\downarrow\downarrow\downarrow\downarrow$ & $\downarrow\downarrow\downarrow\downarrow$ & $\downarrow\downarrow\downarrow\downarrow$\\
\midrule
OP length & $\downarrow\downarrow\downarrow\downarrow$ & $\downarrow\downarrow\downarrow\downarrow$ & $\uparrow\uparrow\uparrow\uparrow$\\
PC length & $\uparrow\uparrow\uparrow\uparrow$ & $\uparrow\uparrow\uparrow\uparrow$ & $\uparrow\uparrow\uparrow\uparrow$\\
Length difference & $\downarrow\downarrow\downarrow\downarrow$ & $\downarrow\downarrow\downarrow\downarrow$ & $\uparrow\uparrow\uparrow\uparrow$\\
Avg. word length difference & $\downarrow\downarrow\downarrow\downarrow$ & $\downarrow\downarrow\downarrow\downarrow$ & $\downarrow\downarrow\downarrow\downarrow$\\
OP/PC POS distribution difference & $\downarrow\downarrow\downarrow\downarrow$ & $\downarrow\downarrow\downarrow\downarrow$ & $\downarrow\downarrow\downarrow\downarrow$\\
Depth of the PC in the thread & $\uparrow\uparrow\uparrow\uparrow$ & $\uparrow\uparrow\uparrow\uparrow$ & $\uparrow\uparrow\uparrow\uparrow$\\
\bottomrule
\end{tabularx}

\caption{Full testing results after Bonferroni correction.}
\label{tab:sig_tests_stop}
\end{table*}

\begin{table*}[t]
\centering
\small

\begin{tabularx}{0.9\textwidth}{Xc}
\toprule
Feature & Total Gain (\%)\\
\midrule
Inverse document frequency & 16.97\\
Stem length & 0.15\\
Wordnet depth (min) & 0.12\\
Wordnet depth (max) & 0.1\\
Stem transfer probability & 46.7\\
\midrule
OP ADP & 0.02\\
OP PRON & 0.1\\
OP X & 0.01\\
OP DET & 0.02\\
OP ADJ & 0.01\\
OP PROPN & 0.02\\
OP VERB & 0.04\\
OP PART & 0.01\\
OP CCONJ & 0.0\\
OP INTJ & 0.01\\
OP NOUN & 0.04\\
OP NUM & 0.01\\
OP ADV & 0.15\\
OP PUNCT & 0.01\\
OP SYM & 0.0\\
OP AUX & 0.0\\
OP subject & 0.53\\
OP object & 0.01\\
OP other & 0.02\\
OP term frequency & 3.23\\
OP normalized term frequency & 0.26\\
OP \# of surface forms & 0.01\\
OP location & 0.15\\
OP in quotes & 0.01\\
OP is entity & 0.02\\
\midrule
PC ADP & 0.02\\
PC PRON & 0.09\\
PC X & 0.81\\
PC DET & 0.05\\
PC ADJ & 0.01\\
PC PROPN & 0.02\\
PC VERB & 0.01\\
PC PART & 0.02\\
PC CCONJ & 0.13\\
PC INTJ & 0.1\\
PC NOUN & 0.04\\
PC NUM & 0.70\\
PC ADV & 0.02\\
PC PUNCT & 0.03\\
PC SYM & 0.2\\
PC AUX & 0.0\\
PC subject & 0.01\\
PC object & 0.01\\
PC other & 0.02\\
PC term frequency & 3.33\\
PC normalized term frequency & 2.92\\
PC \# of surface forms & 0.02\\
PC location & 0.24\\
PC in quotes & 0.04\\
PC is entity & 0.02\\
\midrule
In both OP and PC & 4.88\\
\# of unique surface forms in OP & 0.01\\
\# of unique surface forms in PC & 0.03\\
Stem POS distribution difference & 0.29\\
Stem dependency distribution difference & 0.28\\
\midrule
OP length & 3.63\\
PC length & 3.47\\
Length difference & 2.59\\
Avg. word length difference & 2.65\\
OP/PC POS distribution difference & 3.15\\
Depth of the PC in the thread & 1.4\\
\bottomrule
\end{tabularx}

\caption{Feature importance for the full XGBoost model, as measured by total gain.}
\label{tab:feature_importance}
\end{table*}

\begin{table*}[t]
\small
\centering
\begin{tabular}{p{6cm}p{4cm}r}
\toprule
& Without features & With features \\
\midrule
encoder type & brnn & brnn \\
glove vector dimension & 300 & 300 \\
rnn size & 512 & 512\\
dropout & 0.2 & 0.1 \\
optim & adagrad & adam \\
learning rate & 0.15 & 0.001 \\
beam size & 10 & 10\\
\bottomrule
\end{tabular}
\caption{ Parameters tuned on validation dataset containing 5k instances.}
\label{tab:parameters}
\end{table*}

\subsection{Word--level Prediction Task}
For each non--LSTM classifier, we train 11 models: one full model, and forward and backward models for each of the five feature groups. To train, we fit on the training set and use the validation set for hyperparameter tuning.

For the random model, since the echo rate of the training set is 15\%, we simply predict 1 with 15\% probability, and $0$ otherwise.

For logistic regression, we use the \texttt{lbfgs} solver. To tune hyperparameters, we perform an exhaustive grid search, with $C$ taking values from $\{10^{x}:x\in \{-1, 0, 1, 2, 3, 4\}\}$, and the respective weights of the negative and positive classes taking values from $\{(x, 1-x): x\in\{0.25, 0.20, 0.15\}\}$.

We also train XGBoost models. Here, we use a learning rate of $0.1$, $1000$ estimator trees, and no subsampling. We perform an exhaustive grid search to tune hyperparameters, with the max tree depth equaling 5, 7, or 9, the minimum weight of a child equaling 3, 5, or 7, and the weight of a positive class instance equaling 3, 4, or 5.

Finally, we train two LSTM models, each with a single 300--dimensional hidden layer. Due to efficiency considerations, we eschewed a full search of the parameter space, but experimented with different values of dropout, learning rate, positive class weight, and batch size. We ultimately trained each model for five epochs with a batch size of 32 and a learning rate of 0.001, using the Adam optimizer \citep{kingma2015adam}. We also weight positive instances four times more highly than negative instances.

\subsection{Generating Explanations}
We formulate an abstractive summarization task using an OP concatenated with the PC as a source, and the explanation as target. We train two models, one with the features described above, and one without.
A shared vocabulary of 50k words is constructed from the training set by setting the maximum encoding length to 500 words.
We set the maximum decoding length to 100.
We use a pointer generator network with coverage for generating explanations, using a bidirectional LSTM as an encoder and a unidirectional LSTM as a decoder.
Both use a 256-dimensional hidden state.
The parameters of this network are tuned using a validation set of five thousand instances.
We constrain the batch size to 16 and train the network for 20k steps, using the parameters described in \tableref{tab:parameters}.

\begin{table*}
\small
\centering
\begin{tabular}{p{0.95\textwidth}}
\toprule
\textbf{Original Post}:I keep seeing this point when people bitch about escort quests . But I 've been thinking about it and like , consider the alternatives : 1 ) The NPC moves at your walking speed . Clearly this is a terrible option . Nobody has ever willingly moved at their walking speed in a video game unless they were trying to finesse something or sneak . a walking speed escort quest would be terrible . The fact people even mention this point when talking about NPCs is insane . The actual complaint is " NPC is slower than my run speed " . If the NPC exclusively moved your walk speed it would be 10000 times worse . 2 ) The NPC moves at your run speed . This seems better at first ... but it means that you ca n't pull ahead of the NPC if you want to , or catch up to them if they ever get ahead of you because you stopped to do anything . They 're moving at 100 \% of your max speed . Monsters up ahead ? That 's a fucking shame because you do n't have time to run up and pull aggro on them if the NPC is behind you and you are n't going to be able to intercept them in time at all if the NPC 's ahead because they 'll always get there first . 3 ) The NPC moves at your exact speed behind you following your pathfinding and dynamically navigating traps / moving parts to keep a uniform distance from you . This renders the escort quest pointless . This is a solution to a different problem ( that escort quests are just terrible ) . If having to escort and NPC does n't have any effect on your gameplay decisions they 're a pointless inclusion . The NPC is generally SUPPOSED to require your attention . After all , the only reason you care that they move slower than you is because you have to watch over them . In games where you do n't you just run ahead and let them be slow and it 's no problem . If you need to watch over them , then they need to act in a way where you ca n't just ignore them . Like , if you want to say escort quests are just terrible in general then I 'm on board . Escort quests suck . But if you have to have one with an NPC that has their own movement and pathfinding then they need to move close enough to your run speed that you are n't walking , but far enough from it that you can control your distance from them to some degree while ahead and can catch up when behind . Of all the options available for NPC movement speeds " about 75 - 85 \% of PC run speed " is the best for escort missions both in terms of being least annoying for the player and most able to create the gameplay changes the devs want to create with escort quests .  .  \\
\midrule
\textbf{persuasive comment}: I think the best solution is allowing the player to select a speed equal to the escort . The frustration does n't stem from having to move slower than you would without an escort . It 's that there 's no convenient way to move without running off and leaving them . In real life , it 's simple to adjust your movement speed to a slower person . It 's not about what pace you 're moving , it 's about being able to match pace .\\
\midrule
\textbf{Reference Explanation}: Hmm ... maybe ? I was initially 100\% convinced by \oppc{thinking} on it , I dunno . I feel that the \opword{annoying} thing is just that they 're \oppc{slow}. Like , the fact you have to \oppc{run} laps on them when things are going \opword{well} FEELS \opword{annoying} but I \pcword{think} the \opword{ACTUAL} \opword{annoying} thing is just that it 's \oppc{slow} and because them \oppc{moving} \oppc{slow} causes that you fixate on it . \oppc{Running} at \pcword{pace} with them \oppc{would} be \pcword{equally} \opword{annoying} because you 'd still know you COULD \opword{go} faster ... That said you could be \opword{right} and I 'm convinced \opword{enough} that I \oppc{think} it 's worth a delta \\
\midrule
\textbf{Generated Explanation w/o features}: This is a very good \opword{point} . I had n't thought about it that \oppc{way} . Thank you for your time . I did n't \pcword{think} of it that \oppc{way}\\
\midrule
\textbf{Generated Explanation with features}: I 'm going to give you a delta because you did n't \opword{change} my \opword{view} , but you 've convinced me that there is a \opword{difference} between \oppc{escort} and \oppc{escort} ..\\
\bottomrule
\end{tabular}

\caption{Random generation from Open-NMT Pointer generator network with and without features. In our features model, words like ``escort'' are copied from the OP, but neither model is able to construct a coherent, human-like sentence addressing the explanatory context.}
\end{table*}

\begin{table*}
\small
\centering
\begin{tabular}{p{0.95\textwidth}}
\toprule
\textbf{Original Post}: Hi cmv , This post is not about whether or not abortion is morally permissible or ought to be legal . Rather , it 's a meta - view about the way the abortion debate is structured . Often , those on either side of the debate invoke the circumstances of the pregnancy to support their arguments . Speaking broadly , pro - choice advocates often point to sexual violence or lack of consent as a trump example . Pro - life advocates tend to argue that sex is a responsibility and that women who engage in casual sex are obligated to see a pregnancy through based on that decision . Logically , however , I ca n't see how the circumstances of a pregnancy hold bearing on whether an abortion is morally justifiable . Once a pregnancy has occurred , via any course of action , the moral quandary is the same - does the mother 's right to bodily autonomy take precedence over the fetus ' right to life ? Pick your favorite set of hypothetical circumstances , but at the end of the day the decision at hand is the same , and the logic that brings you to your conclusion ought to apply universally . While I understand the gut instinct to bring up rape and promiscuity when discussing this sensitive issue , I fail to see what bearing they hold to the core question , and believe they only serve to cloud and emotionally charge the debate . I 'm interested in well - structured arguments that connect the circumstances of pregnancy to the abortion debate . CMV ! \\
\midrule
\textbf{persuasive comment}: A major part of the abortion debate on the Pro - Life side is about legitimacy of the action . To someone that is Pro - Life abortion is the killing of a human being , but the killing of a human being is not by default murder . It is only murder if that killing is both unjustifiable and illegal . For those that want abortion to be classified as murder that means that they have to determine if the act is justifiable or not , and for many it is not an all or nothing proposition . Many who are Pro - Life consider abortion due to medical risk to the mother , severe disability of the child , or the child being the result of rape to be legitimate reasons to get an abortion and thus justifiable and in turn action that should be legal . It is the use of an abortion simply because you do not want a child , or because you forgot to use protection that they consider murder . This distinction in kind of human killing is also why killing in war , self defense , defense of another , and an actual pure accident not due to negligence are not crimes as well as why killing as an accident due to negligence is classified as manslaughter and not murder. \\
\midrule
\textbf{Reference Explanation}: To \pcword{someone} that is \oppc{Pro} - \oppc{Life} \oppc{abortion} is the \pcword{killing} of a \pcword{human} being , but the \pcword{killing} of a \pcword{human} being is not by \pcword{default} \pcword{murder} . It is only \pcword{murder} if that \pcword{killing} is both \pcword{unjustifiable} and \pcword{illegal}. For those that \pcword{want} \oppc{abortion} to be \pcword{classified} as \pcword{murder} that \pcword{means} that they have to \pcword{determine} if the \pcword{act} is \oppc{justifiable} or not , and for \pcword{many} it is not an all or \pcword{nothing} \pcword{proposition} . You \pcword{get} a delta for this . The \pcword{distinction} between \opword{ending} a \oppc{life} and \pcword{murder} is valid , and when operating with the latter definition , it makes sense that you 'd have to \pcword{consider} the extraneous factors . I may be limiting the \oppc{debate} too much by interpreting the \oppc{pro} - \oppc{life} position to be against \opword{ending} a \oppc{life} \opword{rather} than against \pcword{murder} .\\
\midrule
\textbf{Generated Explanation w/o features}: ``It is only about \opword{whether} or not \oppc{abortion} is \opword{morally} \opword{permissible} or \opword{ought} to be \opword{legal}''. \opword{Rather} , it 's a \opword{meta} - \opword{view} about the \opword{way} the \opword{decision} at \opword{hand} is the same , and the \pcword{killing} of a \pcword{human} , or the \pcword{child} being the \pcword{result} of \oppc{rape} to be \pcword{legitimate} \pcword{reasons} to \pcword{get} an \oppc{abortion} and \pcword{thus} \oppc{justifiable} and in \pcword{turn} that they have to \pcword{determine} if the \pcword{act} is \opword{morally} . Once , \oppc{pro} - \opword{choice} \opword{advocates} \opword{tend} to \opword{argue} that \opword{sex} is a \opword{responsibility} and that \opword{women} who \opword{engage} in \opword{casual} \opword{sex} are\\
\midrule
\textbf{Generated Explanation with features}: ``It is only \pcword{murder} if that \pcword{killing} is both \pcword{unjustifiable} and \pcword{illegal} . For those that \pcword{want} \oppc{abortion} to be \pcword{classified} as \pcword{murder} that \pcword{means} that they have to \pcword{determine} if the \pcword{act} is \oppc{justifiable} or not , and for \pcword{many} it is not an all or \pcword{nothing} \pcword{proposition}''. !delta. \pcword{Many} who are \oppc{Pro} - \oppc{Life} \pcword{consider} \oppc{abortion} \pcword{due} to \pcword{medical} \pcword{risk} that they \pcword{consider} \pcword{murder} . This \pcword{distinction} in \pcword{kind} of \pcword{human} \pcword{killing} is \pcword{also} why \pcword{killing} as an \pcword{accident} \pcword{due} to the \oppc{mother} , or the \pcword{child} being the \pcword{result} of \oppc{rape} to be \pcword{legitimate}\\
\bottomrule
\end{tabular}

\caption{Random generation from Open-NMT Pointer generator network with and without features. We can observe that the generated explanations transfer entire quotes from the explanandum, indicating extractive summarization capabilities.}
\end{table*}

\begin{table*}
\small
\centering
\begin{tabular}{p{0.95\textwidth}}
\toprule
\textbf{Original Post}: When people use adblockers , they are hurting both consumers and producers . Adblockers take away the primary source of income for websites . Enough people use adblockers that this can seriously jeopardize the finances of a website . These sites include wikis , local newspapers , and many other valuable online resources . If the situation gets bad enough , it forces the producer to do one of 2 things . 1 . Shut down . OR 2 . Move to some sort of paid subscription service . Either way , the world just lost some valuable free information . This hurts the consumers . The benefits of adblockers are small compared to these consequences . Most people justify their use of adblockers by saying they want to avoid viruses / scams and/or intrusive / page - blocking / annoying ads If you are tech savvy enough to get an adblocker , you are probably tech savvy enough to understand what websites you should avoid . Plus you probably have an anti virus anyways . If you 're bothered by intrusive ads , just do n't visit the damn website . Shitty ads are the price you pay for going to some websites . If you are n't willing to pay that price , do n't go to those websites . That simple . That 's all I have to say I guess . i 've just seen too many good websites go down the drain because of this . \\
\midrule
\textbf{persuasive comment}: " Most people justify their use of adblockers by saying they want to avoid viruses / scams and/or intrusive / page - blocking / annoying ads " " If you are tech savvy enough to get an adblocker , you are probably tech savvy enough to understand what websites you should avoid . Plus you probably have an anti virus anyways . " Why avoid the website when you can neuter it with an adblocker ? It still has the content you were looking for after all . You 're also discounting the massive resource savings adblocking can cause . One university deployed an adblocker and saw their traffic go down 30 \% . That 's huge , and that s just network resources , how many cpu cycles get wasted every second running poorly written javascript ads ? How much of your battery goes towards rendering ads ?  \\
\midrule
\textbf{Reference Explanation}: That 's a \opword{good} point about \oppc{resource} \pcword{wasting} . Not sure I 100\% agree that that makes \oppc{adblockers} worth it but I think that 's a valid reason for \oppc{using} \oppc{one} . I 'll give you a delta \\
\midrule
\textbf{Generated Explanation w/o features}: That 's a \opword{good} point . I had n't thought about it that \opword{way} , but I 'll give you a delta for making me realize that it would be better. Thank you for changing my view\\
\midrule
\textbf{Generated Explanation with features}: I 'm \oppc{going} to give you a delta because I did n't really think of it in a \opword{way} that makes sense to me . i 'm just \oppc{going} to give you a delta.\\
\bottomrule
\end{tabular}

\caption{Random generation from Open-NMT Pointer generator network with and without features. Here we see that both the generated examples fail to summarize any concepts specific to the explanandum. Each instead generates a template explanation for a view change.}
\end{table*}

\begin{table*}
\small
\centering
\begin{tabular}{p{0.95\textwidth}}
\toprule
\textbf{Original Post}: People 's main argument is that the poor will have to play money , but they would only have to pay very little because they make so little . It would make everyone feel that they are accomplishmisg something for the nation . Also I am also saying that the rich will also pay their fair ammounts . I forgot where but it was calcuated if everyone would pay a 24 \% tax it would work out for America . Also I would argue for a flat tax because it completely eliminate the need for the Internal Revenue Service . Flat taxes are especially popular in the business and investment communities , where it is argued that because income from dividends , capital gains and distributions is untaxed , freeing up money that would have gone to taxes , investments and savings are thereby encouraged . But that is just my view , so feel free to change it . :) \\
\midrule
\textbf{persuasive comment}: I think it 's worth examining what you mean by fair . My read of your flat tax is that you think equal treatment gives equal / fair footing . Since you mentioned elsewhere that you think minimum wage should not be a permanent career choice , I assume you think people should start at minimum wage but take a better job as soon as possible . So , here are some practical questions : 1 . How is a person supposed to go on job interviews if they are already not making enough money to support them self ? 2 . How is a person supposed to spend money economically if they never have enough of it on a reliable basis to take advantage of sales , bulk purchases , payment plans , etc . ? 3 . How is a person supposed to got to school if they do n't have money to pay for their education ? And what responsible lender would loan someone that money knowing how little income they have available to make payments ? On the other side of the spectrum : 1 . How is a millionaire harmed by paying a higher marginal rate , given that he already has all of his basic need met , and then some ? 2 . What is the benefit to a millionaire putting his money into savings ? Does n't that just make less money available to the rest of the economy ? 3 . Is it ' fair ' to allow a millionaire to invest in and profit in a business , while paying those whose labor produces the profits for that business make less than a living wage ? \\
\midrule
\textbf{Reference Explanation}: \pcword{1} they \pcword{start} with a \pcword{minimum} \pcword{wage} \pcword{job} and \opword{work} themselves up , that s how its always been \pcword{2} Plenty of ppl do nt have \pcword{enough} \oppc{money} right now , I am not \opword{saying} that the system is perfect \pcword{3} There are plenty of \oppc{people} that \pcword{give} \pcword{loans} to students that do nt even have a \pcword{job} 4 He is \pcword{harmed} by losing \oppc{money} ? 5 They will use it to buy things thus boosting the econ . 6 Yea , because that s how they set it up and the gov . \pcword{allows} it to happen If anything boost the min . \pcword{wage} delta\\
\midrule
\textbf{Generated Explanation w/o features}: That 's a good point .Thanks for \opword{changing} my \opword{view}. delta\\
\midrule
\textbf{Generated Explanation with features}: That 's a good point . I was n't aware of it that way . I still \pcword{think} that it should be a \pcword{permanent} \pcword{career} , but that 's a good point .\\
\bottomrule
\end{tabular}

\caption{Random generation from Open-NMT Pointer generator network with and without features.}
\end{table*}

\end{document}